%% file: main.tex
\RequirePackage{fix-cm}
\documentclass[twocolumn]{svjour3}          
\smartqed  
\usepackage{graphicx}
\usepackage{subcaption}
\usepackage{cite}
\usepackage{amsmath,amssymb,amsfonts}
\usepackage{algorithmic}
\usepackage{textcomp}
\usepackage{xcolor}
\usepackage{url}
\usepackage{xcolor, soul}
\sethlcolor{yellow}

\newcommand{\figref}[1]{Fig.~\ref{#1}}
\newcommand{\secref}[1]{Section~\ref{#1}}
\newcommand{\tabref}[1]{Table~\ref{#1}}

\makeatletter
\newcommand*\bigcdot{\mathpalette\bigcdot@{.5}}
\newcommand*\bigcdot@[2]{\mathbin{\vcenter{\hbox{\scalebox{#2}{$\m@th#1\bullet$}}}}}
\makeatother

\begin{document}

\title{EfficientHRNet

}
\subtitle{Efficient and Scalable High-Resolution Networks for \\Real-Time Multi-Person 2D Human Pose Estimation}

\author{Christopher Neff         \and
        Aneri Sheth \and Steven Furgurson \and John Middleton \and Hamed Tabkhi
}

\institute{Christopher Neff \at
              \email{cneff1@uncc.edu}           
           \and
           Aneri Sheth \at
              \email{asheth2@uncc.edu}
           \and
           Steven Furgurson \at
              \email{sfurgurs@uncc.edu}
           \and
           John Middleton \at
              \email{jmiddl11@uncc.edu}
           \and
           Hamed Tabkhi \at
              \email{htabkhiv@uncc.edu}
            \and University of North Carolina at Charlotte, NC, USA
}

\date{Received: date / Accepted: date}

\maketitle

\input{tex/Abstract}

\input{tex/Introduction}

\input{tex/RelatedWork}

\input{tex/Methodology}

\input{tex/Results}

\input{tex/Conclusion}

\input{tex/Acknowlegment}

%
%


\bibliographystyle{spmpsci}
\bibliography{egbib}

\end{document}

%% file: tex/Abstract.tex
\begin{abstract}
There is an increasing demand for lightweight multi-person pose estimation for many emerging smart IoT applications. However, the existing algorithms tend to have large model sizes and intense computational requirements, making them ill-suited for real-time applications and deployment on resource-constrained hardware. Lightweight and real-time approaches are exceedingly rare and come at the cost of inferior accuracy. In this paper, we present EfficientHRNet, a family of lightweight multi-person human pose estimators that are able to perform in real-time on resource-constrained devices. By unifying recent advances in model scaling with high-resolution feature representations, EfficientHRNet creates highly accurate models while reducing computation enough to achieve real-time performance. The largest model is able to come within 4.4\% accuracy of the current state-of-the-art, while having 1/3 the model size and 1/6 the computation, achieving 23 FPS on Nvidia Jetson Xavier. Compared to the top real-time approach, EfficientHRNet increases accuracy by 22\% while achieving similar FPS with $\frac{1}{3}$ the power. At every level, EfficientHRNet proves to be more computationally efficient than other bottom-up 2D human pose estimation approaches, while achieving highly competitive accuracy.
\keywords{Human Pose Estimation \and High-Resolution Networks \and Model Scaling \and Real-Time \and Lightweight}
\end{abstract}

%% file: tex/Introduction.tex
\section{Introduction}
\input{Figures/Intro_Scatter}
Two-dimensional human pose estimation is a common task used in many popular smart applications and has made substantial progress in recent years. There are two primary approaches to 2D human pose estimation. The first is a top-down approach, where cropped images of humans are provided and the network uses those cropped images to produce human keypoints. Top-down approaches rely on object detectors to provide initial human crops, thus they often come with relatively higher computation cost, and are not truly end-to-end. The second is a bottom-up approach, where a network works off the original image and produces human keypoints for all people in the image. While these methods often do not quite reach the accuracy that is possible with state-of-the-art top-down approaches, they come with relatively lower model size and computational overhead. Even so, state-of-the-art bottom-up approaches are still quite large and computationally expensive. The current state-of-the-art \cite{higherhrnet} having 63.8 million parameters and requiring 154.3 billion floating-point operations.

Many emerging Internet-of-Things (IoT) applications require lightweight real-time multi-person pose estimation at the edge, next to the cameras. This is more pronounced in a broad range of smart and connected applications with demands for continuous human activity analysis and behavioral monitoring. Few examples are video surveillance, patient monitoring, and public safety \cite{REVAMP2T,PatientPose,AutonomousDrivingPose}. All these applications demand agile but highly accurate human pose estimation that can run next to the cameras on the IoT edge devices. Despite this, there has been a dearth of attention towards developing lightweight bottom-up methods that are capable of real-time execution under constrained computational resources. To address the gap, there is a need for a family of lightweight real-time human pose estimation models that achieves comparable accuracy to the state-of-the-art approaches.

In this paper, we present EfficientHRNet\footnote{The source code of EfficientHRNet has been provided here: \url{https://github.com/TeCSAR-UNCC/EfficientHRNet}}, a family of lightweight scalable networks for high-resolution and efficient real-time bottom-up multi-person pose estimation. EfficientHRNet unifies the principles of state-of-the-art EfficientNet \cite{efficientnet} and HRNet \cite{hrnet_pose}, and presents a new formulation that enables near state-of-the-art human pose estimation while being more computationally efficient than all other bottom-up methods. Similar to HRNet, EfficientHRNet uses multiple resolutions of features to generate keypoints, but in a much more efficient manner. At the same time, it uses EfficientNet as a backbone and adapts its scaling methodology to be better suited for human pose estimation. To enable lightweight real-time execution, EfficientHRNet further expands the EfficientNet formulation to not only scale below the baseline, but also to jointly scale down the input resolution, High-Resolution Network, and Heatmap Prediction Network. Through this, we are able to create a family of networks that can address the entire domain of real-time 2D human pose estimation while being flexible towards the accuracy and computation requirements of an application.

We evaluate accuracy on the COCO dataset \cite{COCO} and real-time performance on the Nvidia NX Xavier. \figref{fig:ScatterPlot} demonstrates how our models provide equivalent or higher accuracy at lower computational costs than their direct peers. When comparing to state-of-the-art models, baseline EfficientNet competes in accuracy while requiring much less computation, resulting in faster inference. Compared to HRNet \cite{hrnet_pose}, EfficientHRNet achieves 0.4\% higher accuracy while reducing computation requirements by 34\%. When comparing to HigherHRNet \cite{higherhrnet} and PersonLab \cite{personlab}, EfficientHRNet sees between a 1.7\% to 5.1\% decrease in accuracy, while reducing computation requirements by an impressive 83\% to 93\%. This results in a 3.4x FPS increase over HigherHRNet. Even when comparing to models designed specifically for lightweight execution, such as Lightweight OpenPose \cite{lightweight_mobv1}, a scaled-down EfficientHRNet is able to achieve 10.1\% higher accuracy while further reducing computation by 15\%, maintaining similar FPS while requiring $\frac{1}{3}$ the power. In addition, the scaled-down backbone models have been evaluated in isolation on ImageNet. The results demonstrate competitive accuracies while achieving greater efficiency than their peers.

In summary, this article has the following contributions:
\vspace{-3pt}
\begin{itemize}
    \item We propose EfficientHRNet as the first approach to provide lightweight, scalable models for bottom-up real-time multi-person 2D human pose estimation that achieves comparable accuracy to the state-of-the-art.
    \item We propose a novel formulation integrating the scalability of EfficientNet throughout our high-resolution networks to reduce the computational complexity and allow for real-time execution.
    \item We are the first to provide a downwards scaling formulation to create a chain of compact EfficientNet models that scale below the baseline for embedded and edge IoT devices with restrained computation power.
    \item We provide a comprehensive analysis on the challenging COCO dataset \cite{COCO} to show how our models compare against both state-of-the-art and real-time focused approaches in accuracy, model size, computational complexity, efficiency, and real-time execution.
    \item We perform extensive performance analysis on state-of-the-art embedded IoT GPU (Nvidia Jetson NX \cite{NX_Xavier}) to demonstrate the execution benefits of EfficientHRNet over existing algorithms.
    \vspace{-6pt}
\end{itemize}

The rest of this paper continues as follows: \secref{sec:RelatedWork} provides and overview of related work in the field of human pose estimation, high-resolution networks, and scalable neural networks. \secref{sec:EfficientHRNet} details the EfficientHRNet architecture and the novel joint scaling formulation. \secref{sec:Results} provides extensive results and comprehensive analysis on the challenging COCO dataset, as well as comparisons to other state-of-the-art and real-time approaches, and a qualitative assessment of EfficientHRNet's scaling. Finally, \secref{sec:Conclusion} presents concluding remarks.

%% file: Figures/Intro_Scatter.tex
\begin{figure}[b]
    \begin{center}
        \includegraphics[width=1.0\linewidth,trim= 153 470 220 125,clip]{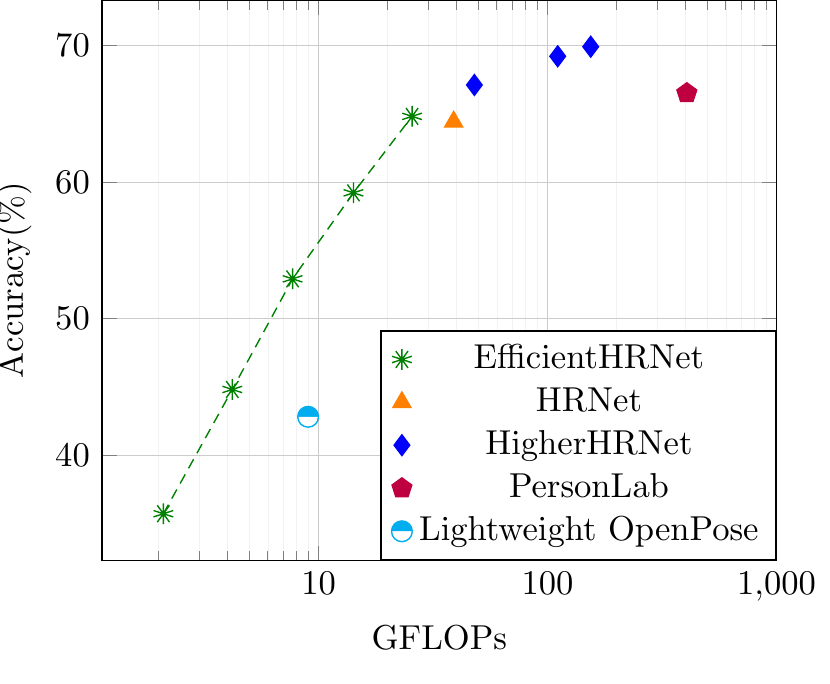}
    \end{center}
    \vspace{-15pt}
    \caption{A comparison of computational complexity and accuracy between bottom-up human pose estimation methods. Accuracy measured on COCO2017 val dataset. X-axis is logarithmic in scale.}
    \label{fig:ScatterPlot}
    \vspace{-10pt}
\end{figure}

%% file: tex/RelatedWork.tex
\section{Related Work}\label{sec:RelatedWork}

\subsection{Top-down Methods}

Top-down methods rely on first identifying all the persons in an image using an object detector, and then detecting the keypoints for a single person within a defined bounding box. 
These single person \cite{td1,td2,td3,td4,td5,td6,td7,stacked_hour_glass,td8,td9,td11} and multi-person \cite{td10,td12,td13,rmpe,td14} pose estimation methods often generate person bounding boxes using object detector \cite{od1,od2,od3,od4}. For instance, RMPE \cite{rmpe} adds symmetric spatial transformer network on top of single person pose estimator stacked hourglass network \cite{stacked_hour_glass} to get high-quality regions from inaccurate bounding boxes and then detects poses using parametric non-maximum suppression.

\subsection{Bottom-up Methods}

Bottom-up methods \cite{openpose2016,openpose2018,pifpaf,associative_embedding,personlab,deepcut,deepercut,multiposenet} detect identity-free keypoints in an image and group them into persons using various keypoints grouping techniques. Methods like \cite{deepcut} and \cite{deepercut} perform grouping by integer linear program and non-maximum suppression. This allows for faster inference times compared to top-down methods with almost similar accuracies. Other methods further improve upon prediction time by using greedy grouping techniques, along with other optimizations, as seen in \cite{openpose2016,openpose2018,pifpaf,associative_embedding,personlab}. For instance, OpenPose \cite{openpose2016,openpose2018} is a multi-stage network where one branch detects keypoints in the form of heatmaps, while the other branch generates Part Affinity Fields that are used to associate keypoints with each other. Grouping is done by calculating the line integral between all keypoints and grouping the pair that has the highest integral. Lightweight OpenPose \cite{lightweight_mobv1} replaces larger backbone with MobileNet \cite{mobilenets} to achieve real-time performance with fewer parameters and FLOPs while compromising on accuracy. PifPaf \cite{pifpaf} uses Part \textit{Intensity} Fields to detect body parts and Part \textit{Associative} Fields for associating parts with each other to form human poses. In \cite{associative_embedding}, a stacked hourglass network \cite{stacked_hour_glass} is used both for predicting heatmaps and grouping keypoints. Grouping is done by assigning each keypoint with an embedding, called a tag, and then associating those keypoints based on the $L_2$ distance between the tag vectors. In this paper, we mainly focus on a highly accurate, end-to-end multi-person pose estimation method as in \cite{associative_embedding}.

\subsection{Multi-scale High-Resolution Networks}  

Feature pyramid networks augmented with multi-scale representations are widely adopted for complex and necessary computer vision applications like segmentation and pose estimation \cite{deeplab,scale-aware_seg,pyramid_det,pyramid_pose,cascade_pyramid}. Recovering high-resolution feature maps using techniques like upsampling, dilated convolution, and deconvolution are also widely popular for object detection \cite{pyramid_det}, semantic segmentation \cite{segnet,unet,deconvnet,exfuse,enc-dec,unified_seg} and pose estimation \cite{stacked_hour_glass,b_conv,multi-scale_struct,pyramid_pose,cascade_pyramid,deepcut,deepercut}. Moreover, there are several works that focus on generating high-resolution feature maps directly \cite{conv_fabrics,interlinked_conv,multi-scale_dense,HRNET,hrnet_pixels,hrnet_pose,higherhrnet}. HRNet \cite{hrnet_pose,hrnet_pixels} proposes to maintain high-resolution feature maps throughout the entire network. HRNet consists of multiple branches with different resolutions across multiple stages. With multi-scale fusion, HRNet is able to generate high resolution feature maps and has found its application in object detection, semantic segmentation, and pose estimation \cite{hrnet_pose,HRNET,hrnet_pixels} thereby achieving remarkable accuracies. Recently, HigherHRNet for multi-person pose estimation \cite{higherhrnet} is proposed which uses HRNet as base network to generate high resolution feature maps, and further adds a deconvolution module to predict accurate, high-quality heatmaps. HigherHRNet achieves state-of-the-art accuracy on the COCO dataset \cite{COCO}, surpassing all existing bottom-up methods. In this paper, we adopt the principles of HigherHRNet \cite{higherhrnet} for generating high-resolution feature maps with multi-scale fusion for predicting high quality heatmaps.

\subsection{Model Scaling}

Previous works on bottom-up pose estimation \cite{openpose2016,openpose2018,higherhrnet,hrnet_pose,associative_embedding,stacked_hour_glass} often rely on either large backbone networks, like ResNet \cite{resnet} or VGGNet \cite{vgg}, or large input resolutions and multi-scale training for achieving state-of-the-art accuracy. Some recent works \cite{hrnet_pose,higherhrnet} show that increasing the channel dimension of otherwise identical models can further improve accuracy. EfficientNet \cite{efficientnet} and RegNet \cite{regnet} show that by jointly scaling network width, depth, and input resolution, better efficiency for image classification can be achieved compared to previous state-of-the-art networks using much larger models. More recently, EfficientNet's lite models remove elements, such as squeeze and excite and swish layers, to make the network more hardware friendly. Inspired by EfficientNet, EfficientDet \cite{efficientdet} proposes a compound scaling method for object detection along with efficient multi-scale feature fusion. We observe that there is a lack of an efficient scaling method for multi-person pose estimation, especially for embedded devices. Lightweight pose estimation models which are scalable and comparatively accurate are needed for computer vision applications which focus on real-time performance. Our proposed compound scaling, also inspired by EfficientNet, is a method that jointly scales the width, depth, and input resolution of EfficientHRNet, as well as the repetition within the \textit{high-resolution modules}, explained in \secref{sec:EfficientHRNet}. In addition, this compound scaling allows our EfficientNet backbone to scale below the baseline B0, creating even lighter weight models.

\subsection{Real-Time Pose Estimation}

While most work in the field focuses on accuracy in isolation, some recent works have been developed that shift the focus more to real-time inference. \cite{Yang2020LightweightDC} focuses on real-time execution, using a densely connected residual module and high-resolution feature maps, similar to \cite{hrnet_pose}, for accurate and lightweight single person pose estimation, able to achieve 39 FPS on an Nvidia 1080TI. \cite{lightweight_mobv1} modifies OpenPose \cite{openpose2018} to use a MobileNet \cite{mobilenets} backbone and fewer refinement stages, and uses the Intel OpenVINO Toolkit \cite{OpenVino} to obtain 28 FPS on an Intel NUC 6i7KYB. Nvidia has also been focusing on real-time inference, releasing trt\_pose \cite{trtpose}, a single person pose estimation model optimized with TensorRT and DeepStream \cite{DeepStream}, achieving up to 251 FPS on the Nvidia Jetson Xavier \cite{NvidiaXavier}.


%% file: tex/Methodology.tex
\begin{figure*}
	\centering
	\includegraphics[width=1\linewidth, trim= 22 19 26 23,clip]{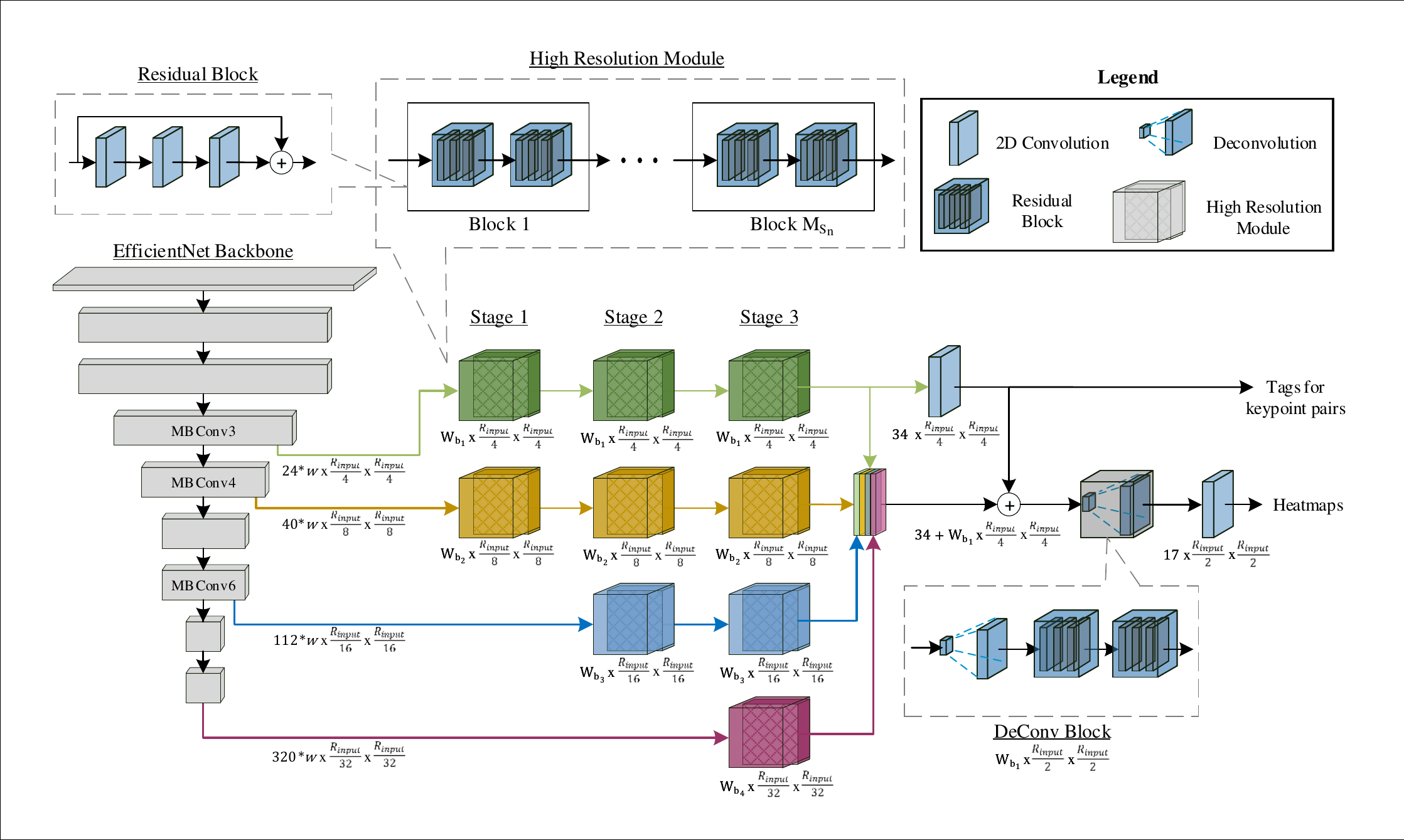}
	\caption{A detailed illustration of the EfficientHRNet architecture. Consisting of a backbone EfficientNet, a High-Resolution Network with three stages and four branches (denoted by different colors), and a Heatmap Prediction Network. EfficinetHRNet is completely scalable, allowing network complexity to be customized for target applications.} \label{fig:Architecture}
\end{figure*}

\section{EfficientHRNet}\label{sec:EfficientHRNet}

We have developed a family of lightweight and scalable networks for real-time multi-person human pose estimation called EfficientHRNet. This section gives an overview of EfficientHRNet and introduces the formulation for the compound scaling of EfficientHRNet's sub-networks.

\subsection{Network Architecture and Formulation}
EfficientHRNet, shown in \figref{fig:Architecture}, comprises of three sub-networks: (1) Backbone Network, (2) High-Resolution Network, and (3) Heatmap Prediction Network. 

\subsubsection{Backbone Network} The first stage of EfficientHRNet is the backbone, consisting of a modified of EfficientNet \cite{efficientnet} altered to scale below the baseline, as discussed in \secref{sec:CompoundScaling}. The backbone outputs four different resolution feature maps of decreasing resolutions $\frac{1}{4}$, $\frac{1}{8}$, $\frac{1}{16}$, and $\frac{1}{32}$ the size of the input image. These feature maps are passed into the main body of the network, called the High-Resolution Network. 

\subsubsection{High-Resolution Network} The High-Resolution Network is inspired by HRNet \cite{hrnet_pose, hrnet_pixels} and HigherHRNet \cite{higherhrnet}. Borrowing the principles of these higher resolution networks brings two major advantages:

\begin{enumerate}
    \item By maintaining multiple high-resolution feature representations throughout the network, heatmaps with a higher degree of spatial precision are generated.
    \item Repeated multi-scale fusions allow for high-resolution feature representations to inform lower-resolution representations, and vice versa, resulting in robust multi-resolution feature representations that are ideal for multi-person pose estimation.
\end{enumerate} 

Figure \ref{fig:Architecture} presents a detailed architecture illustration of EfficientHRNet. It shows the three sub-networks: the Backbone Network, the High-Resolution Network, and the Heatmap Prediction Network. It also provides equations showing how the network scales the input resolution $R_{input}$ and the width of feature maps $W_{b_{n}}$, which will be further explained in \secref{sec:CompoundScaling}.

The High-Resolution Network has three stages $s_1$, $s_2$, and $s_3$ containing four parallel branches $b_1$, $b_2$, $b_3$, and $b_4$ of different resolutions. The first stage $s_1$ starts with two branches $b_1$ and $b_2$, with each consecutive stage adding an additional branch, until all four branches are present in $s_3$. These four branches each consist of \textit{high resolution modules} with a width of $W_{b_{n}}$. Each branch $b_n$ contains feature representations of decreasing resolutions that mirror the resolutions output by the Backbone Network, as shown in \figref{fig:Architecture} and the following equation: 
\input{Equations/branches}

For instance, stage 2 ($s_2$) has three branches of resolutions $\frac{1}{4}$, $\frac{1}{8}$, and $\frac{1}{16}$ of the original input image resolution and a width $W_{b_{n}}$. Moreover, each \textit{high resolution module} is made up of a number of blocks, $M_{s_{n}}$, each containing two \textit{residual blocks}, of which each perform three convolution operations with a residual connection.

\subsubsection{Heatmap Prediction Network} The Heatmap Prediction Network is used to generate human keypoint predictions. In order to predict more accurate heatmaps, a \textit{DeConv block} is added on top of the High-Resolution Network, as proposed in \cite{higherhrnet}. Transposed convolution is used to generate high quality feature maps which are $\frac{1}{2}$ the original input resolution. The input to the \textit{DeConv block} is the concatenation of the feature maps and predicted heatmaps from the High-Resolution Network, as shown in the equation below: 
\input{Equations/concat} 
Two \textit{residual blocks} are added after the deconvolution to refine the up-sampled feature maps. After the \textit{DeConv block}, $1$x$1$ convolution is used to predict heatmaps and tagmaps in a similar fashion to \cite{associative_embedding}, the feature map size of each shown below: \input{Equations/heatmaps_tags}

The grouping process clusters keypoints into multiple persons by grouping keypoints whose tags have minimum $L_2$ distance. Like \cite{higherhrnet}, the High-Resolution Network is scale-aware and uses multi-resolution supervision for heatmaps during training to allow the network to learn with more precision, even for small-scale persons. From the ground truth, heatmaps for different resolutions are generated to match the predicted keypoints of different scales. Thus, the final heatmaps loss is the sum of mean squared errors for all resolutions. However, as high resolutions tagmaps do not converge well, tagmaps are trained on a resolution $\frac{1}{4}$ of the original input resolution, as in \cite{associative_embedding}.

\subsection{Compound Scaling Method} \label{sec:CompoundScaling}
This section details the compound scaling methodology, which jointly scales all parts of EfficientHRNet, as seen in \figref{fig:Architecture} and \tabref{tab:scaling_config}. The aim of EfficientHRNet is to provide a family of models optimized for both accuracy and efficiency, which can be scaled to meet a diverse set of memory and compute constraints.
\begin{table*}[t]
    \centering
    \begin{tabular}{c|c|c|c|c|c|c}
        Model & Input Size & Backbone & Width per Branch &  Blocks per Stage & Tags & Heatmaps \\
     & ($R_{input}$) & Network &  ($W_{b_{1}}$, $W_{b_{2}}$, $W_{b_{3}}$, $W_{b_{4}}$)  & ($M_{s_{2}}$, $M_{s_{3}}$, $M_{s_{4}}$)  &  ($T_{size}$)  &  ($H_{size}$)\\ 
                \hline
                \hline  
            H$_0$ ($\phi$ = 0)      & 512 & $B0$     & 32, 64, 128, 256 & 1, 4, 3 & 128 & 256 \\
            H$_{-1}$ ($\phi$ = -1)  & 480 & $B_{-1}$ & 26, 52, 103, 206 & 1, 3, 3 & 120 & 240 \\ 
            H$_{-2}$ ($\phi$ = -2)  & 448 & $B_{-2}$ & 21, 42, 83, 166  & 1, 2, 3 & 112 & 224 \\
            H$_{-3}$ ($\phi$ = -3)  & 416 & $B_{-3}$ & 17, 34, 67, 133  & 1, 1, 3 & 104 & 208 \\    
            H$_{-4}$ ($\phi$ = -4)  & 384 & $B_{-4}$ & 14, 27, 54, 107  & 1, 1, 2 & 96  & 192 \\
    \end{tabular}
\vspace{+10pt}
    \caption{Efficient scaling configs for EfficientHRNet}
    \label{tab:scaling_config}
    \vspace{-20pt}
\end{table*}
\vspace{-1pt}

Previous works on bottom-up human pose estimation and semantic segmentation mostly scale the base network by using bigger backbone networks like ResNet \cite{resnet} and VGGNet \cite{vgg}, using large input image sizes, or using multi-scale training to achieve high accuracies. However, these methods rely on scaling only a single dimension, which has limited effectiveness. Recent works \cite{efficientnet, regnet} show notable performance on image classification by jointly scaling the width, depth, and input image resolution. Inspired by EfficientNet, EfficientDet \cite{efficientdet} proposes a similar compound scaling method for object detection, which jointly scales the backbone network, multi-scale feature network, and the object detector network. We propose a heuristic-based compound scaling methodology for computer vision applications, specifically bottom-up human pose estimation and semantic segmentation, using EfficientHRNet. Based on \cite{efficientnet}, EfficientHRNet's methodology uses a scaling coefficient $\phi$ to jointly scale the Backbone Network, the High-Resolution Network, and Task-Specific Head. More precisely, the EfficientNet backbone is scaled below the baseline and the rest of EfficientHRNet is scaled down in order to maintain near state-of-the-art accuracy while creating lightweight and flexible networks.

\subsubsection{Backbone Network.} The same width and depth scaling coefficients are maintained as in EfficientNet \cite{efficientnet}. In order to meet the demands of running models on constrained devices, a new formulation for scaling EfficientNet below the baseline and into a more compact model is provided. 

Starting with the baseline EfficientNet-B0 scaling coefficients:
\input{Equations/scalingCoeff}
$\phi$, i.e. $\phi$ = -1, -2, -3, -4, is inverted to calculate the scaling multipliers for the compact EfficientNet models, which is symbolized as $B_{-1}$, $B_{-2}$, $B_{-3}$ and $B_{-4}$ respectively. As an example, in order to take the baseline resolution, 224, and scale it down for our $B_{-1}$ model, we would take $r$, from \eqref{eq:scalingCoeff}, with $\phi = -1$. This would result in a resolution scaling coefficient of $1.15^{-1}$, i.e. 0.87, leaving a scaled resolution size of $ceil(224 * 0.87) = 195$. This pattern repeats for $B_{-2}$ through $B_{-4}$, and can be seen in \tabref{tab:effnet_scaling}. We train these compact EfficientNet models ($B_{-1}$ to $B_{-4}$) on ImageNet and use the resulting models for the Backbone Network in EfficientHRNet.

\subsubsection{High-Resolution Network.} The High-Resolution Network has three stages and four branches with four different feature map sizes. Each branch $n$ also has a different width $W_{b_{n}}$ and our baseline $H_0$ model has a width of 32, 64, 128, and 256 for each branch respectively. We selectively pick a width scaling factor of 1.25 and scale down the width using the following equation: \input{Equations/width_scaling.tex}
where $n$ is a particular branch number and $\phi$ is the compound scaling coefficient. 

Furthermore, within each stage, each \textit{high resolution module} has multiple \textit{blocks} $M_{s_{n}}$ which repeat a number of times, as seen in \tabref{tab:scaling_config}. In our baseline EfficientHRNet $H_{0}$ model, \textit{blocks} within each stage repeat 1, 4, and 3 times respectively. We found that the number of repetitions in stage 3 had the largest impact on accuracy. Therefore, the number of repetitions within a \textit{high resolution module} $M_{s_{2}}$ decreases linearly as the models are scaled down, starting with stage 2 until reaching a single repetition and then moving on to stage 3, as shown in \tabref{tab:scaling_config}.

\subsubsection{Heatmap Prediction Network} The \textit{DeConv block} is scaled in the same manner as the width of the High Resolution Network \eqref{eq:width}. The Heatmap Prediction Network outputs tags and heatmaps whose width remains fixed across all models. 


\subsubsection{Input Image Resolution.} The EfficientNet layers downsample the original input image resolution by 32 times. Thus, the input resolution of EfficientHRNet must be dividable by 32, and is linearly scaled down as shown in the following equation:
\input{Equations/input_scaling}
The final result of this compound scaling methodology on EfficientHRNet $H_{0}$ to $H_{-4}$ can be seen in \tabref{tab:scaling_config}.

%% file: Equations/branches.tex
\begin{equation}
\label{eq:branches}
W_{b_{n}} \times \frac{R_{input}}{2^{n} + 1}
\end{equation}

%% file: Equations/concat.tex
\begin{equation}
\label{eq:concat}
34 + W_{b_{1}} \times \frac{R_{input}}{4} \times \frac{R_{input}}{4}
\end{equation}

%% file: Equations/heatmaps_tags.tex
\begin{equation}
\begin{split}
\label{eq:heatmaps}
T_{size} = 34 \times \frac{R_{input}}{4} \times \frac{R_{input}}{4} \\
H_{size} = 17 \times \frac{R_{input}}{2} \times \frac{R_{input}}{2}
\end{split}
\end{equation}

%% file: Equations/scalingCoeff.tex
\begin{equation}
\label{eq:scalingCoeff}
\begin{split}
depth: d = 1.2^{\phi} \\
width: w = 1.1^{\phi} \\
resolution: r = 1.15^{\phi} \\
\end{split}
\end{equation}

%% file: Equations/width_scaling.tex
\begin{equation}
\label{eq:width}
    W_{b_{n}} = (n \cdot 32)\cdot (1.25)^{\phi} 
\end{equation}

%% file: Equations/input_scaling.tex
\begin{equation}
\label{eq:input_scaling}
    R_{input} = 512 + 32\cdot \phi
\end{equation}

%% file: tex/Results.tex
\section{Experimental Results} \label{sec:Results}
This section evaluates our method for scaling EfficientNet below the baseline through classification on the popular ImageNet \cite{imagenet} and CIFAR-100 \cite{cifar-100} datasets. Then, an exhaustive evaluation of five different EfficientHRNet models is conducted on the challenging COCO \cite{COCO} dataset and compared to state-of-the-art methods. Additional, metrics on real-time inference are reported using the Nvidia Jetson NX Xavier and compared to state-of-the-art lightwieght approaches. Finally, a qualitative evaluation of EfficientHRNet is presented, illustrating both where the models excel and where they fall short.

\subsection{Classification for Compact EfficientNet}

\subsubsection{Dataset}
ImageNet \cite{imagenet} has been a long time standard benchmark for object classification and detection thanks to its annual contest, the ImageNet Large Scale Visual Recognition Challenge, that debuted in 2010. The challenge uses a subset of the full dataset with over a million images spread out over 1000 object classes. For training, validating, and testing purposes, the trimmed ImageNet is divided into three sets: 800k images will be used for training the network, 150k will be used for validation after each epoch, and 50k will be used for testing the fully trained model.
CIFAR-100 \cite{cifar-100} consists of 100 object classes each with 500 images for training, and 100 for testing. This relatively small dataset helps illuminate our lightweight models, which start to struggle with the larger ImageNet as $\phi$ decreases, designed for resource constrained devices that might not need to classify as many object classes.

\subsubsection{Training}
We use random rotation, random scale, and random aspect ratio to crop the input images to the desired resolutions based on the current EfficientNet model. Color jitter was also used to randomly change the brightness, contrast, saturation, and hue of the RGB channels using principle component analysis \cite{alexnet}. The images are then normalized using per channel mean and standard deviation. Each model was trained using Stochastic Gradient Descent \cite{sgd} with a weight decay of $1e-4$. The weights were initialized using the Xavier algorithm \cite{xavier} and underwent five warm-up epochs with a learning rate of $1e-4$ that increased linearly until it reached 0.05. The networks were then trained for an additional 195 epochs and followed the step decay learning rate scheduler \cite{stepdecay} that reduces the learning rate by a factor of 10 every 30 epochs.

\subsubsection{Testing}
The compact EfficientNet models were tested for accuracy based on their respective test sets. For a fair comparison, the number of ImageNet test samples were reduced to 10,000 to match the test set of CIFAR-100, where the batch size is set to 1. These results can be seen in \tabref{tab:effnet_scaling}. 
\input{Tables/EffnetScaling}

\subsubsection{Results on ImageNet and CIFAR-100}
Looking at $B_{-1}$ there is a 15\% reduction in parameters and 25\% reduction in operations, yet an accuracy drop of only 1.2\% and 0.5\% on ImageNet and CIFAR-100 respectively. More impressively, $B_{-2}$ sees a 35-40\% reduction in parameters and a 50\% reduction in operations, yet only a 3.7\% and 2.1\% drop in accuracy on the two datasets. This minor accuracy loss is negligible compared to the massive reduction in model size and computation, allowing for much faster inference as well as deployment on low-power and resource constrained devices. In the most extreme, $B_{-4}$ shows a parameter reduction of 68-75\% and a 87.5\% decrease in operations while having an accuracy drop of 9.4\% and 7.6\% on ImageNet and CIFAR-100. While the accuracy drop is a bit more significant here, the massive reduction in computation allows for much more flexibility when it comes to deployment in systems where a lightweight approach is needed. This gives us a solid foundation on which to build EfficientHRNet.

\subsection{2D Human Pose Estimation for EfficientHRNet}

\subsubsection{Dataset} COCO \cite{COCO} consists of over 200k images with 250k person instances, each annotated with 17 keypoints. COCO is divided into three sets, \textit{train}, \textit{val}, and \textit{test}, which have 57k, 5k, and 40k images respectively. Additionaly, \textit{test-dev} is a subset of \textit{test} with 20k images and is used for fair comparison with other works, where possible. COCO evaluation meterics use mean average precision (AP) and are detailed on the COCO website\footnote{\url{http://cocodataset.org/\#keypoints-eval}}.

\subsubsection{Training} We use random rotation, random scale, and random translation for data augmentation. Following \cite{higherhrnet}, we generate two ground truth heatmaps of different sizes, $\frac{1}{2}$ and $\frac{1}{4}$ of the original input size respectively. Each EfficientHRNet model is trained using Adam optimizer \cite{Adam} and weight decay of $1e-4$. All models from $H_{0}$ to $H_{-4}$ are trained for a total of 300 epochs with a base learning rate of $1e-3$, decreasing to $1e-4$ and $1e-5$ at $200^{th}$ and $260^{th}$ epochs respectively. To maintain balance between heatmap loss and grouping loss, we weight the losses at $1$ and $1e-3$ respectively.

\subsubsection{Testing} Models are tested using both single scale and multi-scale heatmaps, as is common. Following \cite{associative_embedding}, the output detection heatmaps across different scales are averaged and the tags concatenated into higher dimension tags, making the models considerably more scale-invariant.

\input{Tables/Comparison}

\subsubsection{Results on COCO2017 \textit{test-dev}} \tabref{tab:cocotest-dev} compares EfficientHRNet with other bottom-up pose estimation methods on COCO \textit{test-dev} set. The baseline $H_0$ model with single-scale testing serves as an efficient and accurate model for bottom-up methods as it is almost comparable to HRNet \cite{higherhrnet} in accuracy, losing by only 0.1\%, while having 18\% less parameter and 34\% fewer FLOPs. $H_0$ outperforms Hourglass \cite{stacked_hour_glass} in both single-scale and multi-scale testing by 7.4\% and 1.6\% respectively, with $H_0$ remarkably having about $\frac{1}{10}$ the model size and number of FLOPs as Hourglass. The highest performing of all models on COCO \textit{test-dev}, HigherHRNet \cite{higherhrnet}, beats $H_0$ in accuracy by 4.4\%, but at the cost of nearly triple the model size and more than 6x the computation. In all cases where $H_0$ loses in accuracy, it more than makes up for it in a reduction in parameters and operations. Additionally, our $H_{-1}$ model, with only 16M parameters and 14.2B FLOPs, outperforms both OpenPose \cite{openpose2016,openpose2018} and Hourglass \cite{stacked_hour_glass}, demonstrating EfficientHRNet's efficiency and suitability for low-power and resource constrained devices. 

As EfficientHRNet is scaled down using the compound scaling method mentioned in \secref{sec:CompoundScaling}, we see somewhat minor drops in accuracy with significant drops parameters and FLOPs as compared to the baseline $H_{0}$ model. $H_{-1}$ has 31.3\% less parameters and 44.5\% less FLOPs as compared to $H_0$ while only being 4.9\% less accurate. Similarly, our lightest model $H_{-4}$ is 84\% smaller and has 91.7\% less FLOPs, with a less than 45\% drop in accuracy. Interestingly, EfficientHRNet is the only bottom-up pose estimator that is able to provide such lightweight models while still having accuracies that are comparable to state-of-the-art bottom-up methods, as illustrated by both \tabref{tab:cocotest-dev} and \figref{fig:ScatterPlot}. These results nicely show the validity of our approach to scalability and efficiency in EfficientHRNet.

\subsubsection{Results on COCO2017 \textit{val}} We report EfficientHRNet accuracy on COCO \textit{val}, noting the number of parameters and FLOPs, and compare it with other bottom-up methods. In addition, to accurately assess suitability for real-time performance on embedded devices, we inference our models as well as one of our closest competitors, HigherHRnet \cite{higherhrnet}, on the Jetson NX Xavier, first converting the models to ONNX and then inferencing in TensorRT. FPS results for Lightweight OpenPose are on an Intel NUC 6i7KYB as reported in \cite{lightweight_mobv1}. As summarized in Table \ref{tab:cocoval}, the baseline $H_0$ model outperforms HRNet \cite{hrnet_pose} with 0.4\% more accuracy, 18\% fewer parameters and 34\% fewer FLOPs. $H_{-2}$ and $H_{-3}$ models outperform Lightweight OpenPose \cite{lightweight_mobv1} in accuracy while having fewer FLOPs. $H_{-4}$ has the worst accuracy of any model in Table \ref{tab:cocoval}. However, it boasts both the smallest model size and fewest number of operations, seeing an over 75\% reduction from its lightest weight competitor. When looking at FPS, HigherHRNet becomes much less desirable, being the only model unable to achieve at least 20 FPS. $H_0$ is 3.4x faster while only being 2.3\% less accurate. Comparing to Lightweight OpenPose, $H_0$ is 22\% more accurate while only being 2 FPS slower. Scaling down to $H_{-3}$ reduces EfficientHRNet's accuracy lead by to only 2\%, but increases throughput to be 1.3x greater than Lightweight OpenPose. Our smallest model $H_{-4}$ achieves an impressive 50 FPS, but at a substantial cost in accuracy. Note that we see an unusually high FPS drop in $H_{-1}$. This is due to $H_{-1}$'s input resolution and intermediate feature map sizes resulting in memory tiles that map poorly to the NX Xavier's Tensor Core architecture. In the following subsection, we provide further analysis and comparison regarding real-time execution on NX Xavier.

\subsection{Real-Time Execution Analysis on Edge}
Since real-time inference is highly dependent on the hardware utilized, we must account for more than just accuracy and throughput in our comparisons. To best compare accuracy and efficiency across differing platforms, we adopt the Accuracy$\bigcdot$Efficiency (\AE) metric from \cite{REVAMP2T}. \AE~is simply the product of accuracy (measured in AP) and efficiency (measured in FPS per Watt). \tabref{tab:AE} shows how EfficientHRNet compares when taking this information into account. Note that Lightweight OpenPose reports their results on Intel NUC 6i7KYB, which has a TDP of 45 Watts \cite{lightweight_mobv1}, while all other methods were measured on the Jetson NX Xavier, which has a maximum power draw of 15 Watts. We use these power numbers across all approaches in an attempt to create as fair a comparison as possible. EfficientHRNet greatly outperforms the competition in terms of \AE, with all models achieving an \AE~score of over 80 while Lightweight OpenPose and HigherHRNet only achieve scores of 25 and 30 respectively. 

In terms of \AE, EfficientHRNet outperforms the competition between 3x to 5x. This is largely due to the poor throughput of HigherHRNet and the relatively higher power NUC that Lightweight OpenPose reports on. HigherHRNet excels in accuracy and Lightweight OpenPose excels in FPS and model size, while EfficientHRNet is more equally balanced between accuracy, model size, throughput, and power consumption. This gives EfficientHRNet a leg up in terms of low-power, real-time inference, making its scalable models the new state-of-the-art for lightweight bottom-up human pose estimation for real-time edge applications.

\input{Tables/COCOval_results}


\input{Tables/AE}

\subsection{Qualitative Analysis}
\input{Figures/Pose_Qual}
To further demonstrate how EfficientHRNet models perform in relation to one another, we present qualitative results on the COCO dataset. \figref{fig:pose_qual} shows simple, medium, and complex examples for all EfficientHRNet models from $H_0$ to $H_{-4}$. Here, we can see that $H_0$ can accurately detect all but the most distant and occluded individuals. This is accuracy that is functionally identical to state-of-the-art models but is able to inference in real-time, making it immensely valuable for applications the require high accuracy but need to run in real-time or on low-power devices. Looking at $H_{-1}$ we can see that keypoints are accurately detected, but in medium and complex scenarios keypoint groupings become confused. Here, the confusion is minor enough that it can be filtered out with additional post processing, meaning that applications that require predicting complex scenarios on devices that can not fit $H-0$ can use $H_{-1}$ with slightly decreased accuracy. For the medium scenario, there is also missed detection from the distant person occluded by the left-most surfboard, though such missed detections are relatively uncommon throughout the dataset. However, for simple scenarios there is little to no difference when compared with $H_0$. These qualities make $H_{-1}$ a compelling model when using a device without enough memory resources for $H_0$ and when a minor amount of error is acceptable, or for applications that will only deal with simple scenarios. $H_{-2}$ looks a lot like $H_{-1}$, but the confusion is worse, with multiple keypoint grouping being detected for a single person, and this even extends to simple scenarios. This would again require additional post processing, depending on the application. $H_{-3}$ and $H_{-4}$ follow the same pattern, with confusion continuing to get worse. Again, we see that actual keypoint detections themselves are fairly accurate, and the greatly reduced model size and computational complexity open up a wide range of additional devices capable of real-time performance. This makes the smaller models extremely compelling for real-time applications that can afford a certain amount of error but require the use of highly resource constrained devices, particularly in the case of simple scenarios. This analysis helps visualize the relationship between detected keypoints and model size, and shows the overall affect on accurate human pose prediction as we move towards smaller models. This further validates EfficientHRNet as a family of high accuracy and efficient models capable of real-time 2D human pose estimation for a variety of embedded and resource constrained devices.

%% file: Tables/EffnetScaling.tex
\begin{table}[t]
    \centering
    \resizebox{\linewidth}{!}{
    \begin{tabular}{c|c|c|c|c|c|c}
     & \multicolumn{1}{c|}{Input} & & \multicolumn{2}{c|}{ImageNet} & \multicolumn{2}{c}{CIFAR-100} \\
    Model & size & FLOPs & Params & Top-1 & Params & Top-1 \\ 
    \hline \hline
    B0   & 224 & 0.4B & 5.3M & 75  & 4.1M & 81.9 \\ 
    $B_{-1}$  & 195 & 0.3B & 4.5M & 73.8 & 3.5M & 81.4  \\
    $B_{-2}$  & 170 & 0.2B & 3.4M & 71.3 & 2.5M & 79.8  \\
    $B_{-3}$ & 145 & 0.1B & 2.8M & 68.5 & 1.9M & 78.2  \\
    $B_{-4}$  & 128 & 0.05B & 1.3M & 65.6 & 1.3M & 74.3 \\
    \end{tabular}
    }
    \caption{Compact EfficientNet performance on ImageNet and CIFAR-100 datasets.}
    \label{tab:effnet_scaling}
    \vspace{-10pt}
\end{table}

%% file: Tables/Comparison.tex
\begin{table}[!bt]
    \centering
    \resizebox{\linewidth}{!}{
        \begin{tabular}{c|c|c|c|c|c}
                            &                   & Input &           &           &   \\
            Method         & Backbone        & size & \# Params & FLOPs & AP   \\
            \hline \hline                                                                              
            \multicolumn{6}{c}{w/o multi-scale test}                                                                  \\
            \hline                                          
            OpenPose        & -             & -     & 25.94M    & 160B      & 61.8  \\
            Hourglass       & Hourglass     & 512   & 277.8M    & 206.9B    & 56.6  \\
            PersonLab       & ResNet-152    & 1401  & 68.7M     & 405.5B    & 66.5  \\
            PifPaf          & ResNet-152    & -     & -         & -         & 66.7  \\
            
            HRNet           & HRNet-W32     & 512   & 28.5M     & 38.9B     & 64.1  \\
            HigherHRNet     & HRNet-W32     & 512   & 28.6M     & 47.9B     & 66.4  \\
            HigherHRNet     & HRNet-W48     & 640   & 63.8M     & 154.3B    & 68.4  \\
            \hline
            H$_{0}$         & $B0$          & 512   & 23.3M     & 25.6B     & 64.0  \\
            H$_{-1}$        & $B_{-1}$      & 480   & 16M       & 14.2B     & 59.1  \\
            H$_{-2}$        & $B_{-2}$      & 448   & 10.3M     & 7.7B      & 52.8  \\
            H$_{-3}$        & $B_{-3}$      & 416   & 6.9M      & 4.2B      & 44.5  \\
            H$_{-4}$        & $B_{-4}$      & 384   & 3.7M      & 2.1B      & 35.5  \\
        
            \hline                                                                                                       
            \multicolumn{6}{c}{w/ multi-scale test}                                                                   \\
            \hline                                                                                                       
            Hourglass      & Hourglass      & 512   & 277.8M    & 206.9B    & 63.0  \\
            Hourglass      & Hourglass      & 512   & 277.8M    & 206.9B    & 65.5  \\
            PersonLab      & ResNet-152     & 1401  & 68.7M     & 405.5B    & 68.7  \\
            HigherHRNet    & HRNet-W48      & 640   & 63.8M     & 154.3B    & 70.5  \\
            \hline
            H$_{0}$        & $B0$           & 512   & 23.3M     & 25.6B     & 67.1  \\
            H$_{-1}$       & $B_{-1}$       & 480   & 16M       & 14.2B     & 62.3  \\
            H$_{-2}$       & $B_{-2}$       & 448   & 10.3M     & 7.7B      & 55.0  \\
            H$_{-3}$       & $B_{-3}$       & 416   & 6.9M      & 4.2B      & 45.5  \\
            H$_{-4}$       & $B_{-4}$       & 384   & 3.7M      & 2.1B      & 39.7  \\
        \end{tabular}
    }
    \caption{Comparisons with state-of-the-art bottom-up methods on COCO2017 test-dev dataset. Numbers for HRNet come from a bottom-up approach outlined in \cite{higherhrnet}.}
    \label{tab:cocotest-dev}
\end{table}

%% file: Tables/COCOval_results.tex
\begin{table}[t]
    \small
    \resizebox{\linewidth}{!}{
    \begin{tabular}{c|c|c|c|c|c}
    & Input & & & & \\
    Model       &  size        & AP & \# Params & FLOPs & FPS\\
    \hline \hline
    PersonLab               & 1401  & 66.5  & 68.7M & 405.5B & - \\  
    HRNet                   & 512   & 64.4  & 28.5M & 38.9B  & - \\
    HigherHRNet             & 512   & 67.1  & 28.6M & 47.9B  & 6.68 \\ 
    Lightweight OpenPose    & 368   & 42.8  & 4.1M  & 9.0B   & 26 \\ \hline \hline
    H$_0$ ($\phi$ = 0)      & 512   & 64.8  & 23.3M & 25.6B  & 22.95 \\ 
    H$_{-1}$ ($\phi$ = -1)  & 480   & 59.2  & 16M   & 14.2B  & 20.43 \\ 
    H$_{-2}$ ($\phi$ = -2)  & 448   & 52.9  & 10.3M & 7.7B   & 24.53 \\ 
    H$_{-3}$ ($\phi$ = -3)  & 416   & 44.8  & 6.9M  & 4.2B   & 33.78 \\
    H$_{-4}$ ($\phi$ = -4)  & 384   & 35.7  & 3.7M  & 2.1B   & 50.96 \\ 
    
    \end{tabular}
    }
    \caption{Comparisons with bottom-up methods on COCO2017 val dataset. Numbers for HRNet come from a bottom-up approach outlined in \cite{higherhrnet}. FPS results were preformed on the Nvidia Jetson NX Xavier \cite{NX_Xavier}. Lightweight OpenPose numbers were reported on the Intel NUC 6i7KYB.}
    \label{tab:cocoval}
\end{table}

%% file: Tables/AE.tex
\begin{table}[t]
    \small
    \resizebox{\linewidth}{!}{
    \begin{tabular}{c|c|c|c|c|c}
    
    Model                   & AP    & FPS   & Efficiency & \AE \\
    \hline \hline
    HigherHRNet             & 67.1  & 6.68  & 0.445  & 29.850  \\ 
    Lightweight OpenPose    & 42.8  & 26    & 0.578  & 24.738  \\ \hline \hline
    H$_0$ ($\phi$ = 0)      & 64.8  & 22.95 & 1.530  & 99.144  \\ 
    H$_{-1}$ ($\phi$ = -1)  & 59.2  & 20.43 & 1.362  & 80.630  \\ 
    H$_{-2}$ ($\phi$ = -2)  & 52.9  & 24.53 & 1.635  & 86.492  \\ 
    H$_{-3}$ ($\phi$ = -3)  & 44.8  & 33.78 & 2.252  & 100.89  \\
    H$_{-4}$ ($\phi$ = -4)  & 35.7  & 50.96 & 3.397  & 121.273 \\ 

    \end{tabular}
    }
    \caption{\AE~comparisons with lightweight bottom-up approaches. Lightweight OpenPose results reported in Intel NUC 6i7KYB (45W). All other results on the Nvidia Jetson NX Xavier (15W).}
    \label{tab:AE}
\end{table}

%% file: Figures/Pose_Qual.tex
\begin{figure*}[t]
    \begin{minipage}[t]{0.15\textwidth}
        \centering
        \subcaption{$H_0$}
        \includegraphics[width=2.6cm, height=2.15cm]{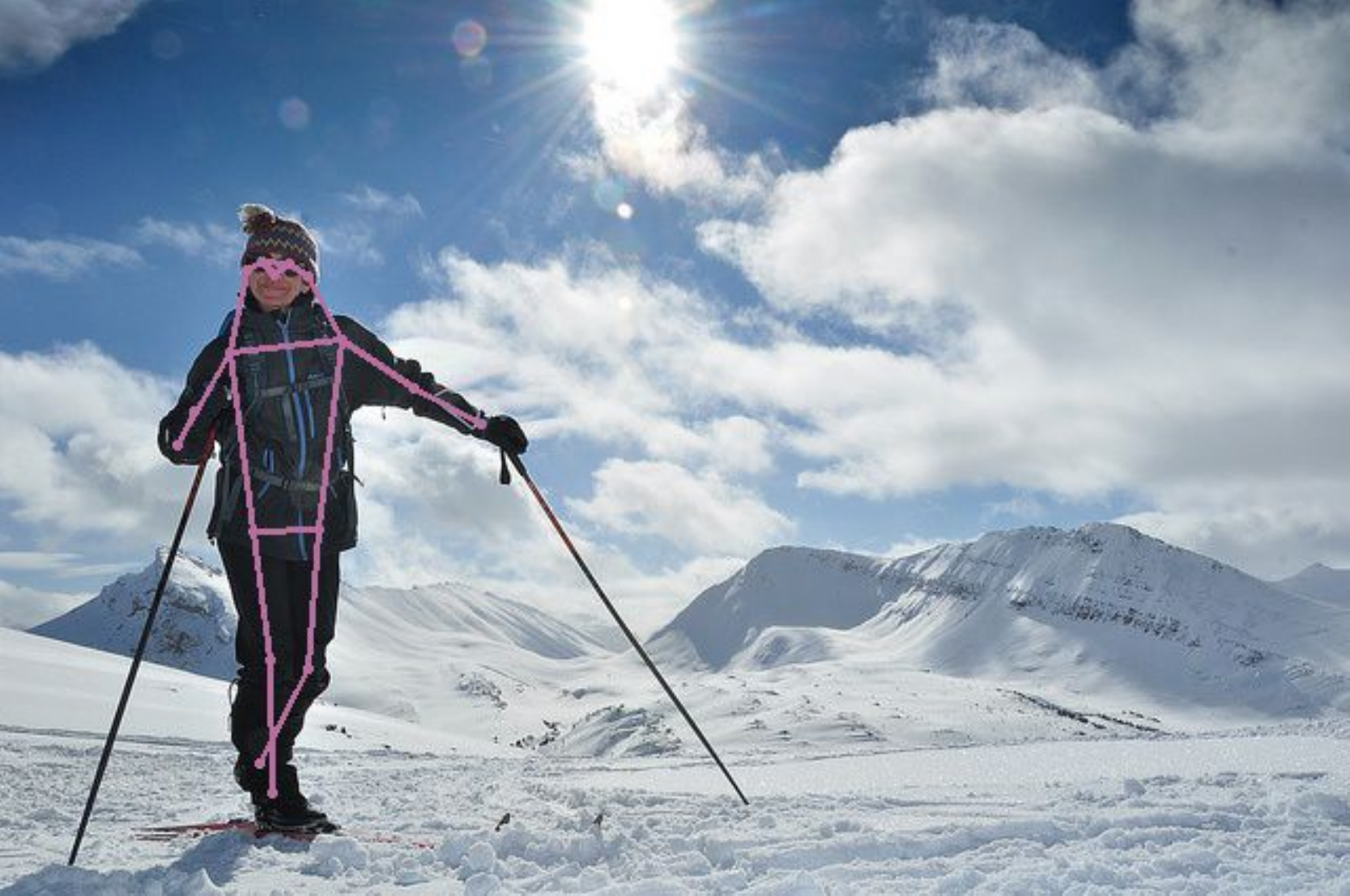}
    \end{minipage}
    \hfill
    \begin{minipage}[t]{0.15\textwidth}
        \centering
        \subcaption{$H_{-1}$}
        \includegraphics[width=2.6cm, height=2.15cm]{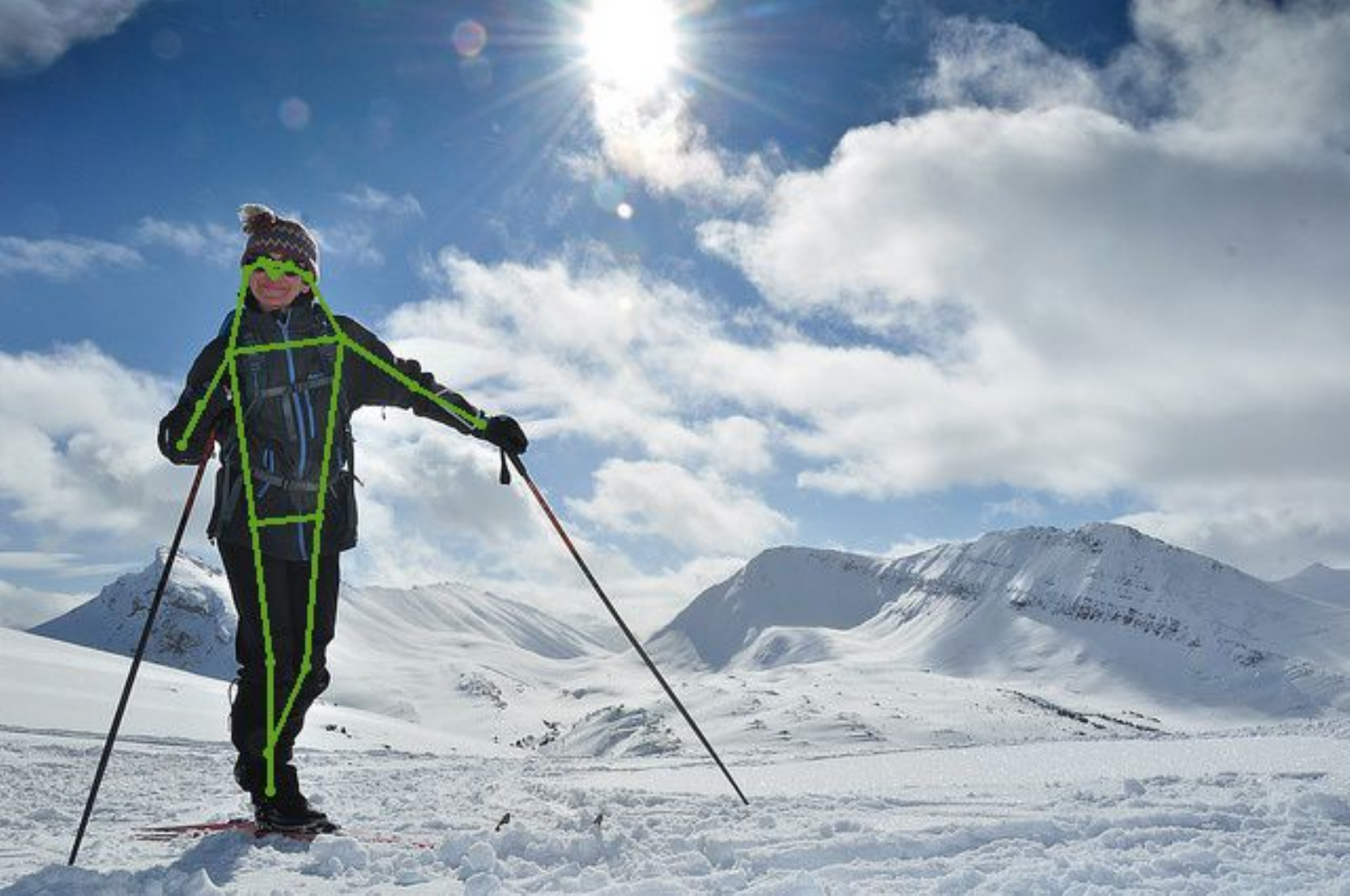}
    \end{minipage}
    \hfill
    \begin{minipage}[t]{0.15\textwidth}
        \centering
        \subcaption{$H_{-2}$}
        \includegraphics[width=2.6cm, height=2.15cm]{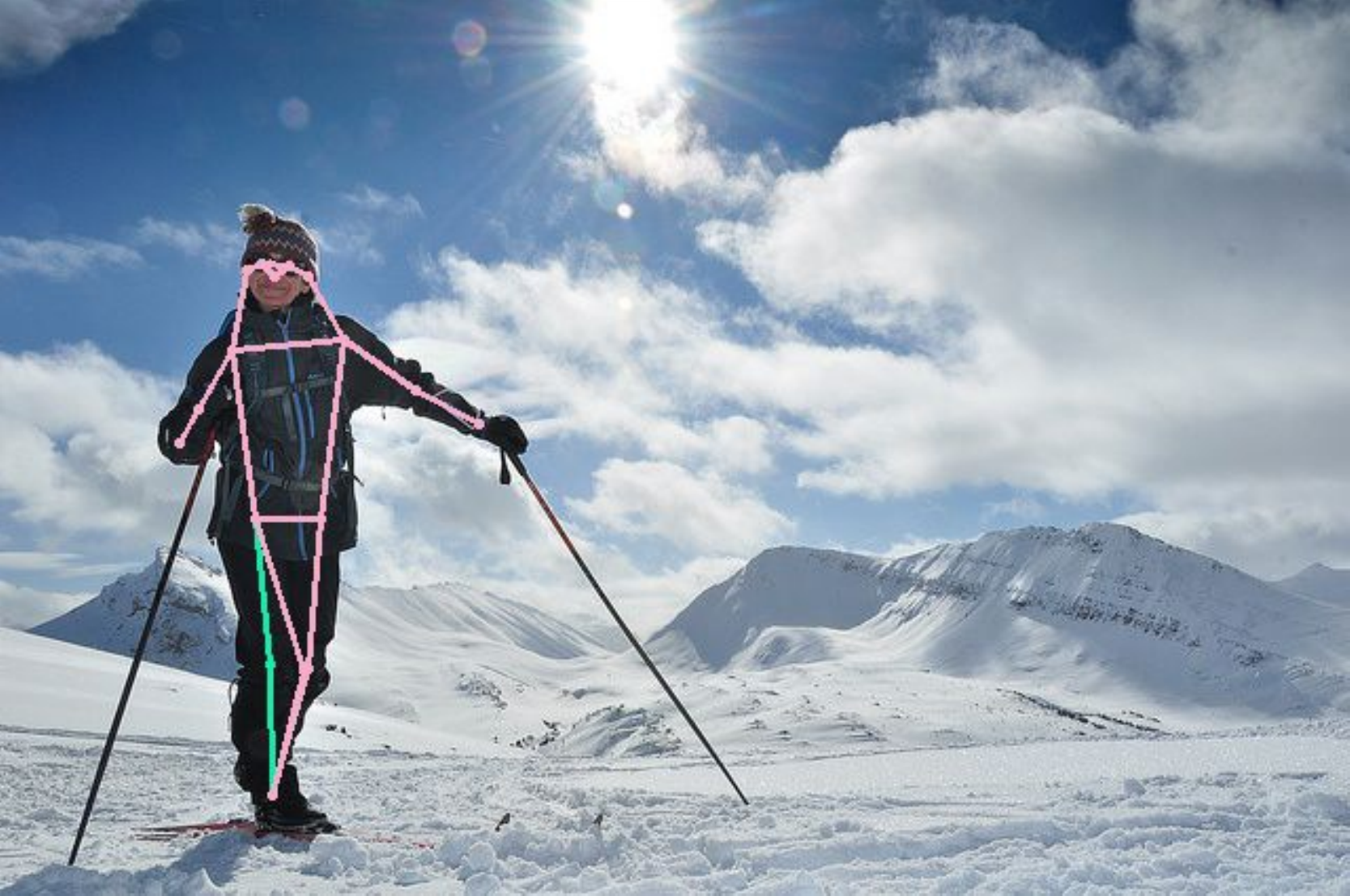}
    \end{minipage}  
    \hfill
    \begin{minipage}[t]{0.15\textwidth}
        \centering
        \subcaption{$H_{-3}$}
        \includegraphics[width=2.6cm, height=2.15cm]{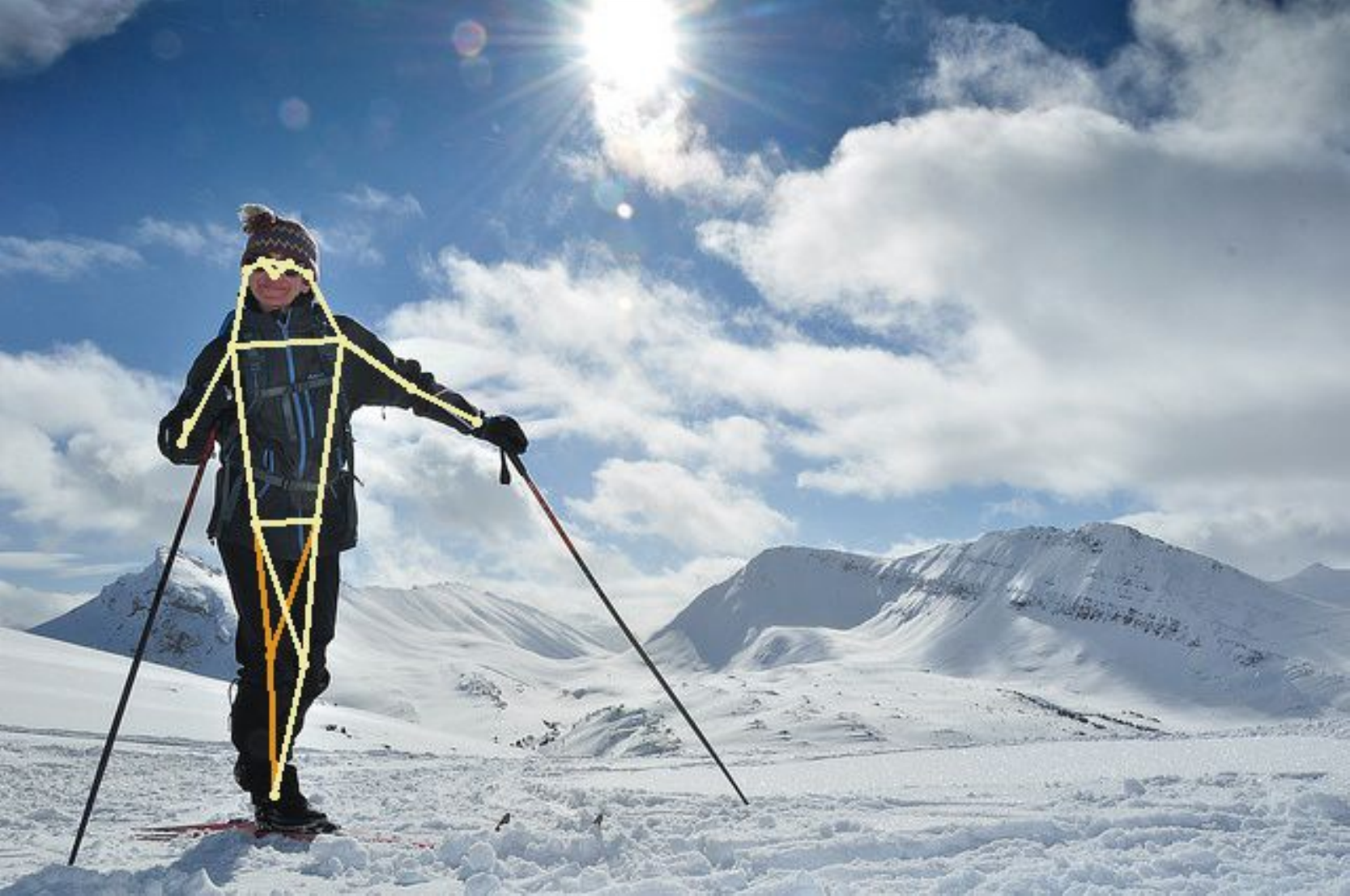}
    \end{minipage}
    \hfill
    \begin{minipage}[t]{0.15\textwidth}
        \centering
        \subcaption{$H_{-4}$}
        \includegraphics[width=2.6cm, height=2.15cm]{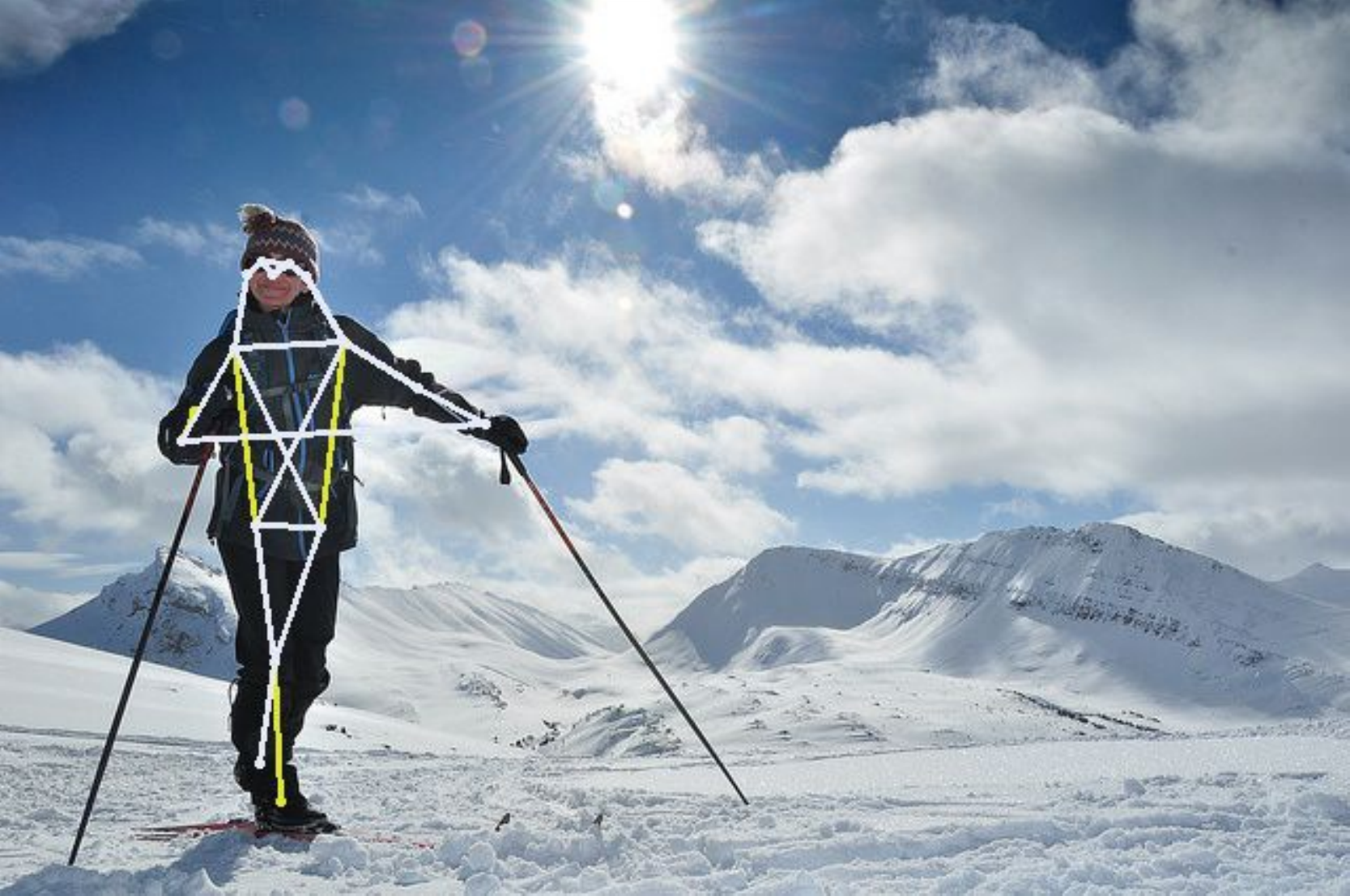}
    \end{minipage}  
    \label{fig:1-2-3}
    \vspace{0.18cm}
    \\
    \begin{minipage}[t]{0.15\textwidth}
        \centering
        \includegraphics[width=2.6cm, height=2.15cm]{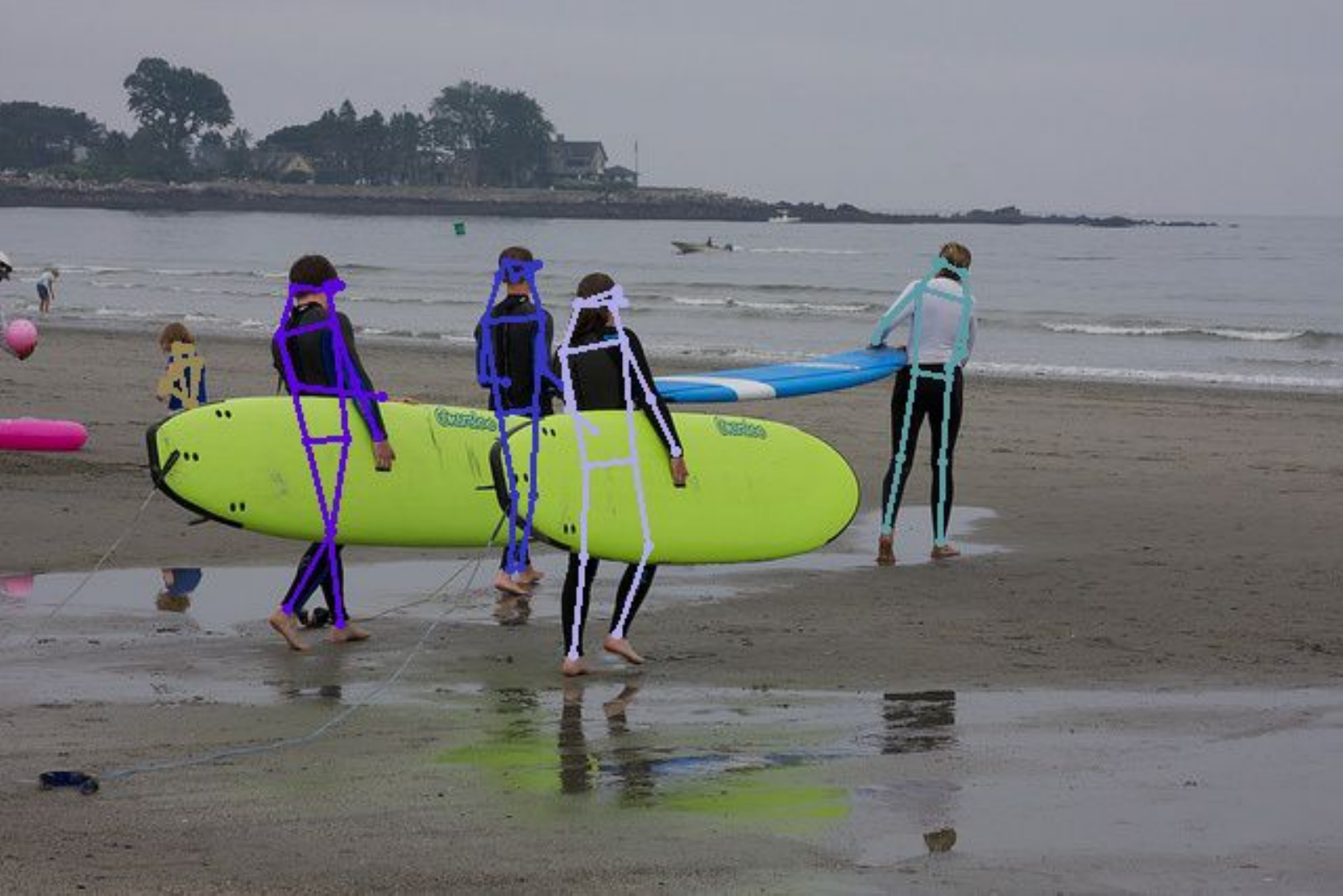}
    \end{minipage}
    \hfill
    \begin{minipage}[t]{0.15\textwidth}
        \centering
        \includegraphics[width=2.6cm, height=2.15cm]{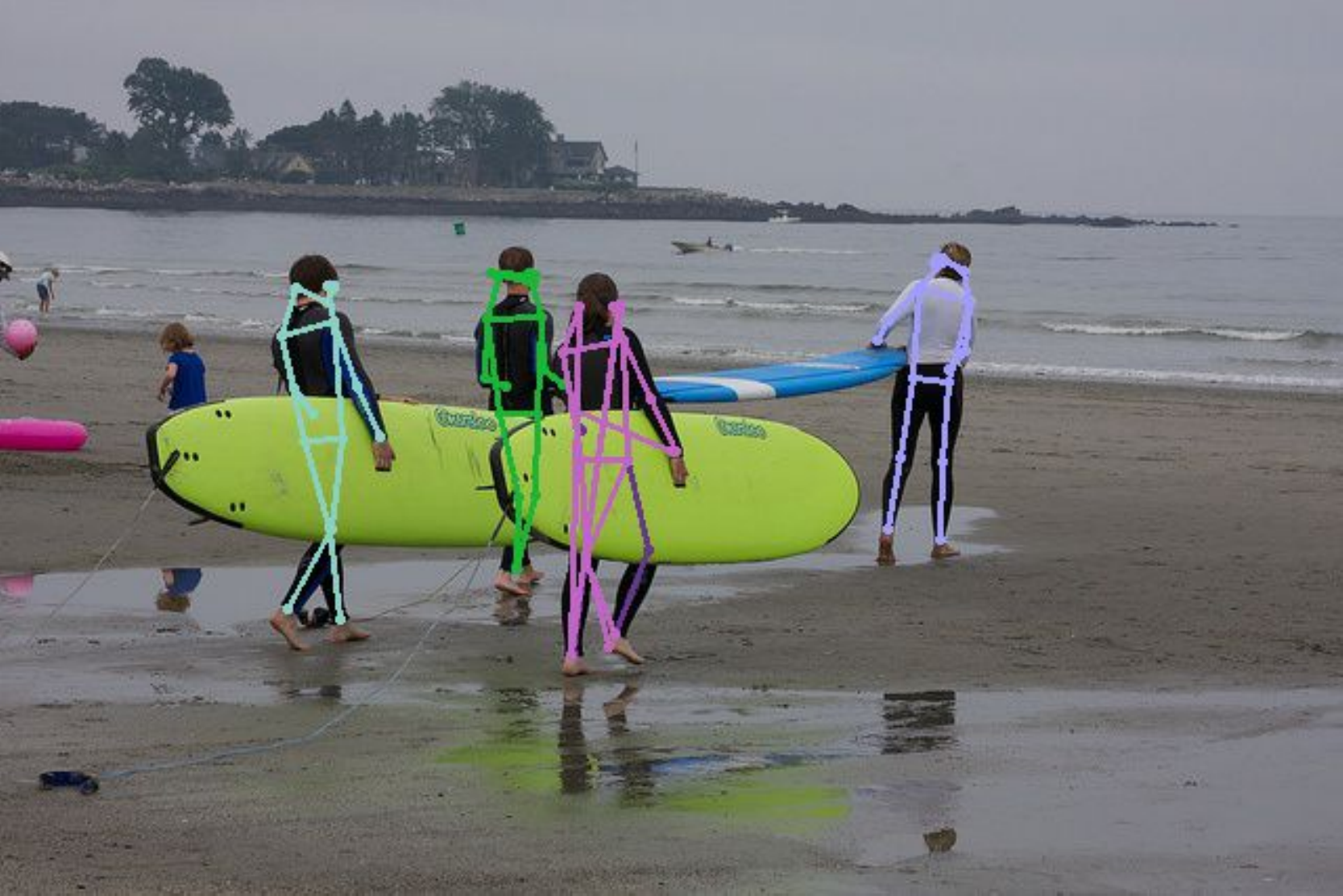}
    \end{minipage}
    \hfill
    \begin{minipage}[t]{0.15\textwidth}
        \centering
        \includegraphics[width=2.6cm, height=2.15cm]{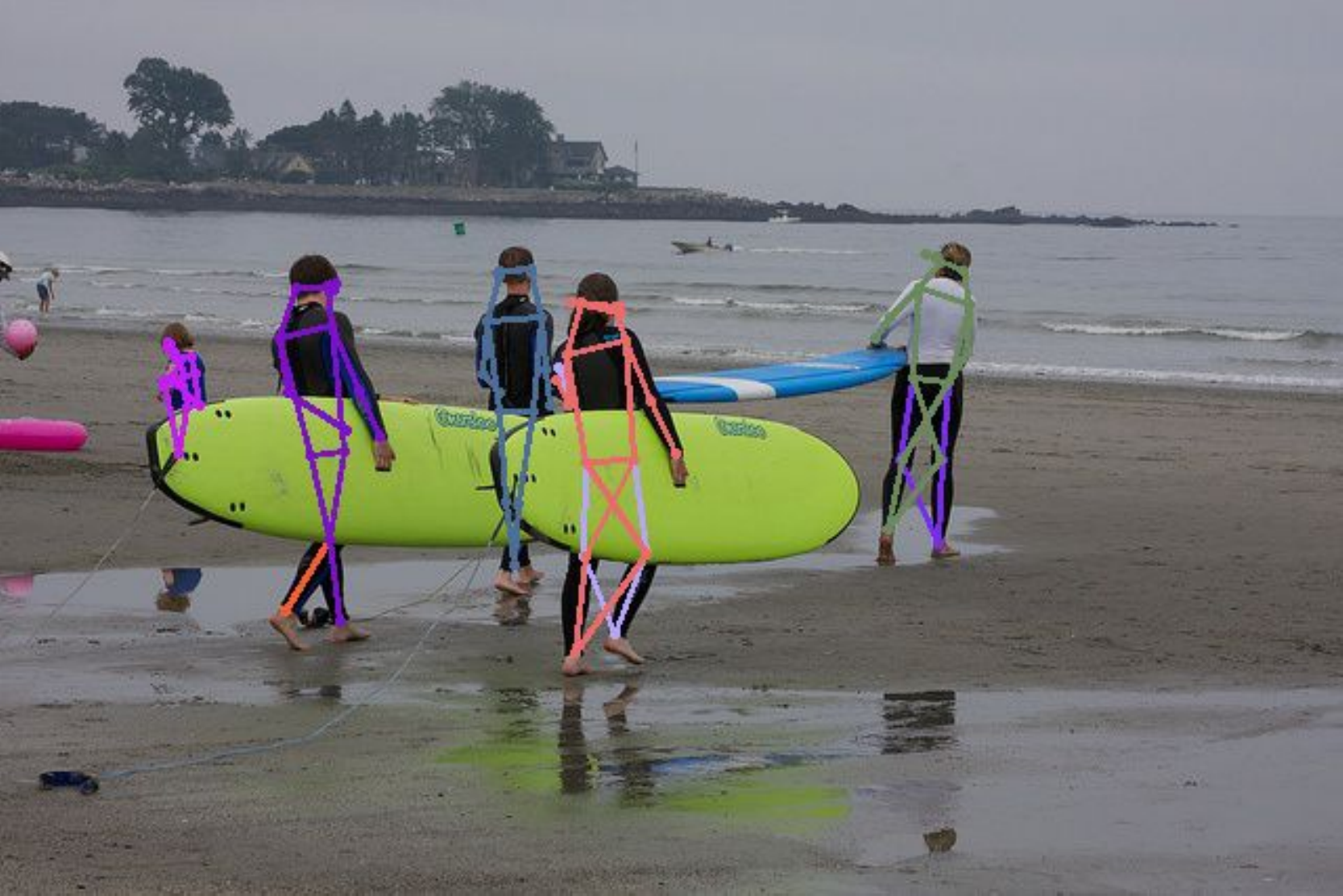}
    \end{minipage}  
    \hfill
    \begin{minipage}[t]{0.15\textwidth}
        \centering
        \includegraphics[width=2.6cm, height=2.15cm]{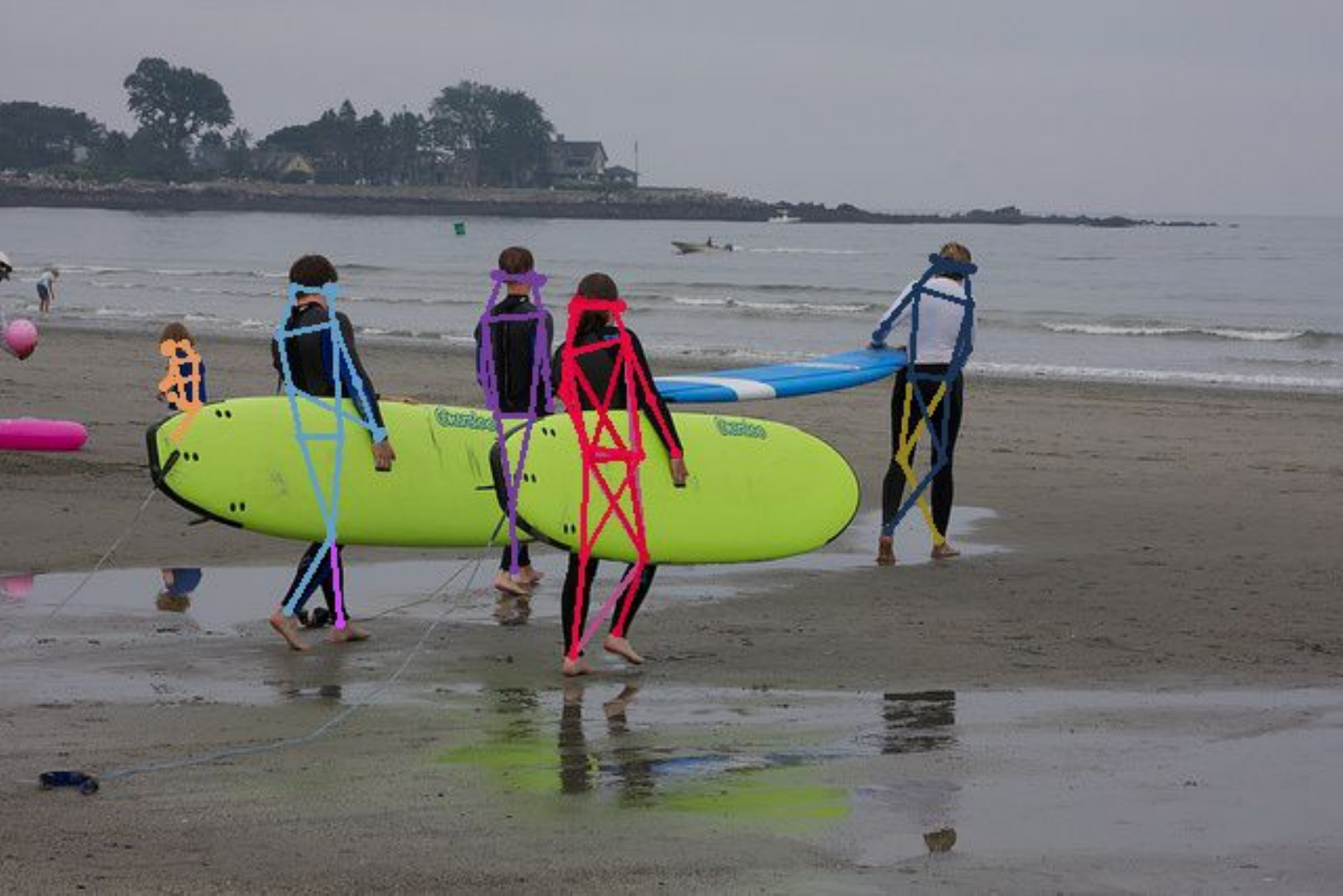}
    \end{minipage}
    \hfill
    \begin{minipage}[t]{0.15\textwidth}
        \centering
        \includegraphics[width=2.6cm, height=2.15cm]{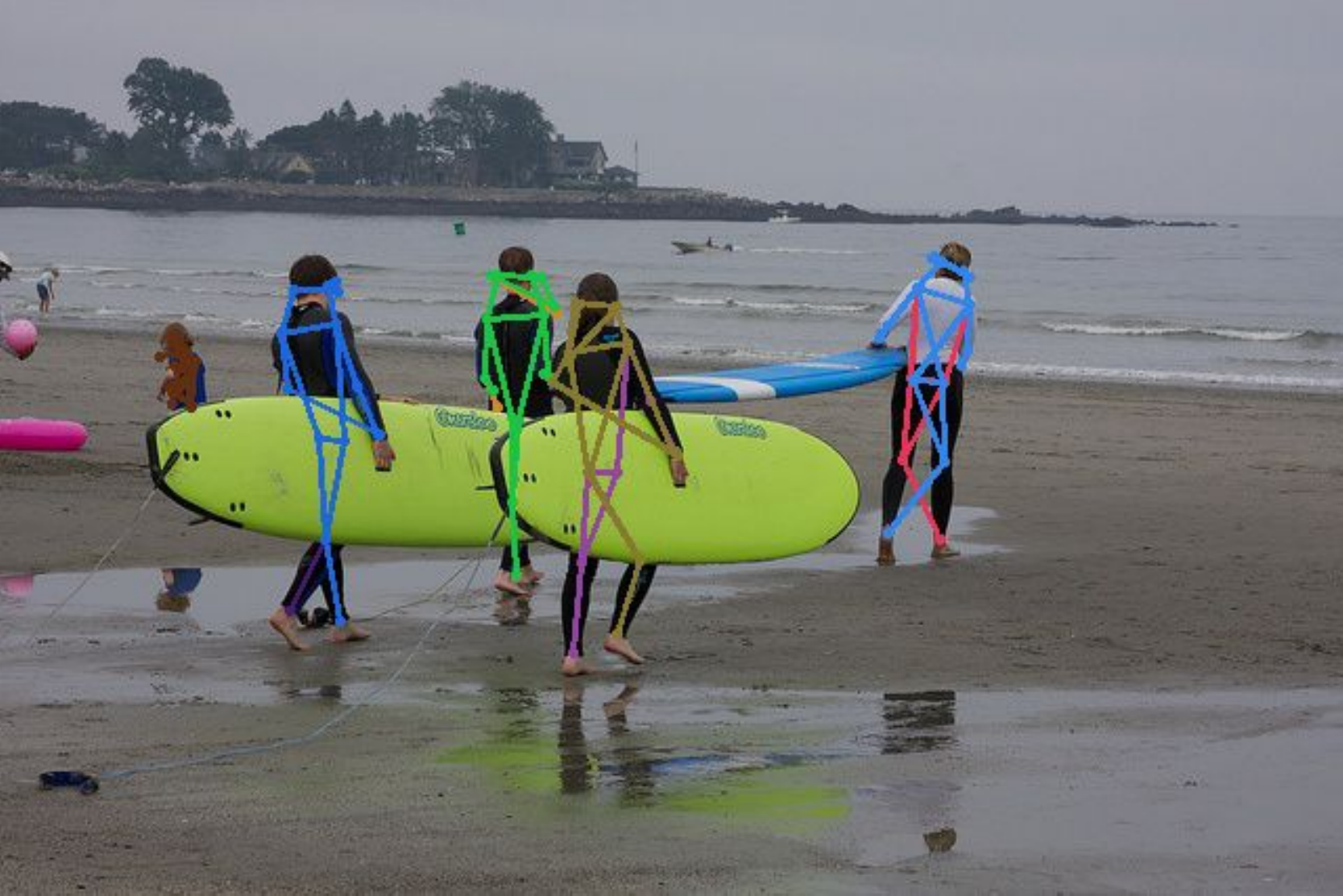}
    \end{minipage}  
    \label{fig:1-2-32}
    \vspace{0.18cm}
    \\
    \begin{minipage}[t]{0.15\textwidth}
        \centering
        \includegraphics[width=2.6cm, height=2.15cm]{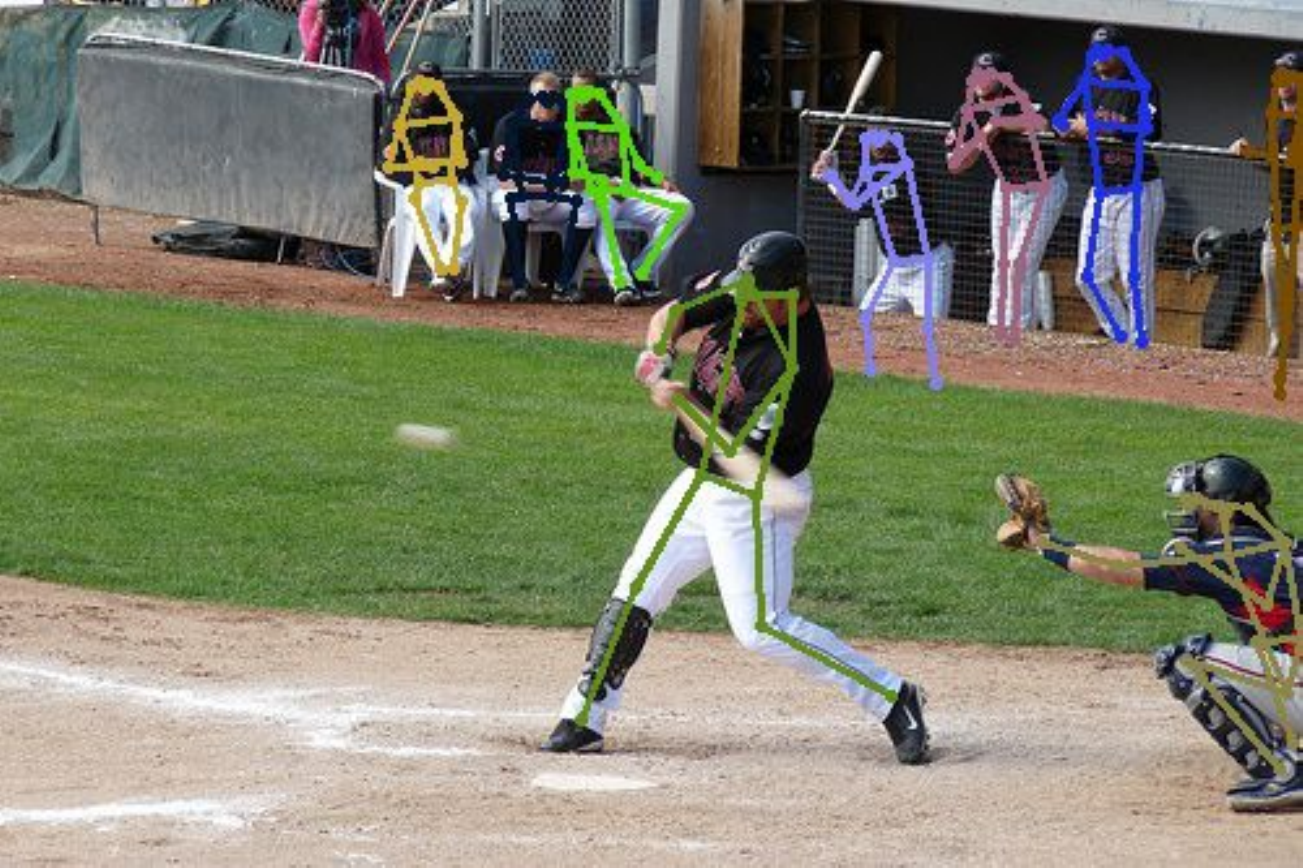}
    \end{minipage}
    \hfill
    \begin{minipage}[t]{0.15\textwidth}
        \centering
        \includegraphics[width=2.6cm, height=2.15cm]{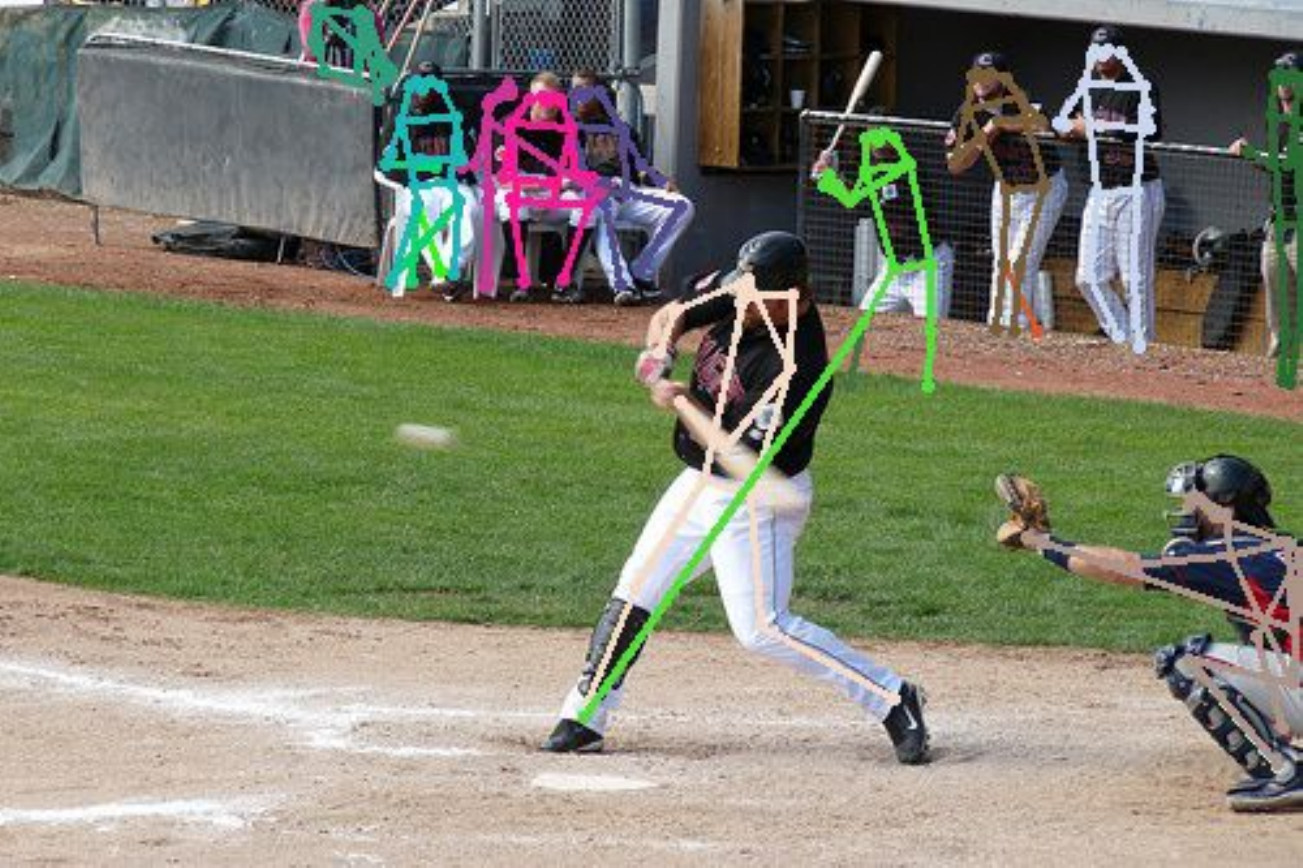}
    \end{minipage}
    \hfill
    \begin{minipage}[t]{0.15\textwidth}
        \centering
        \includegraphics[width=2.6cm, height=2.15cm]{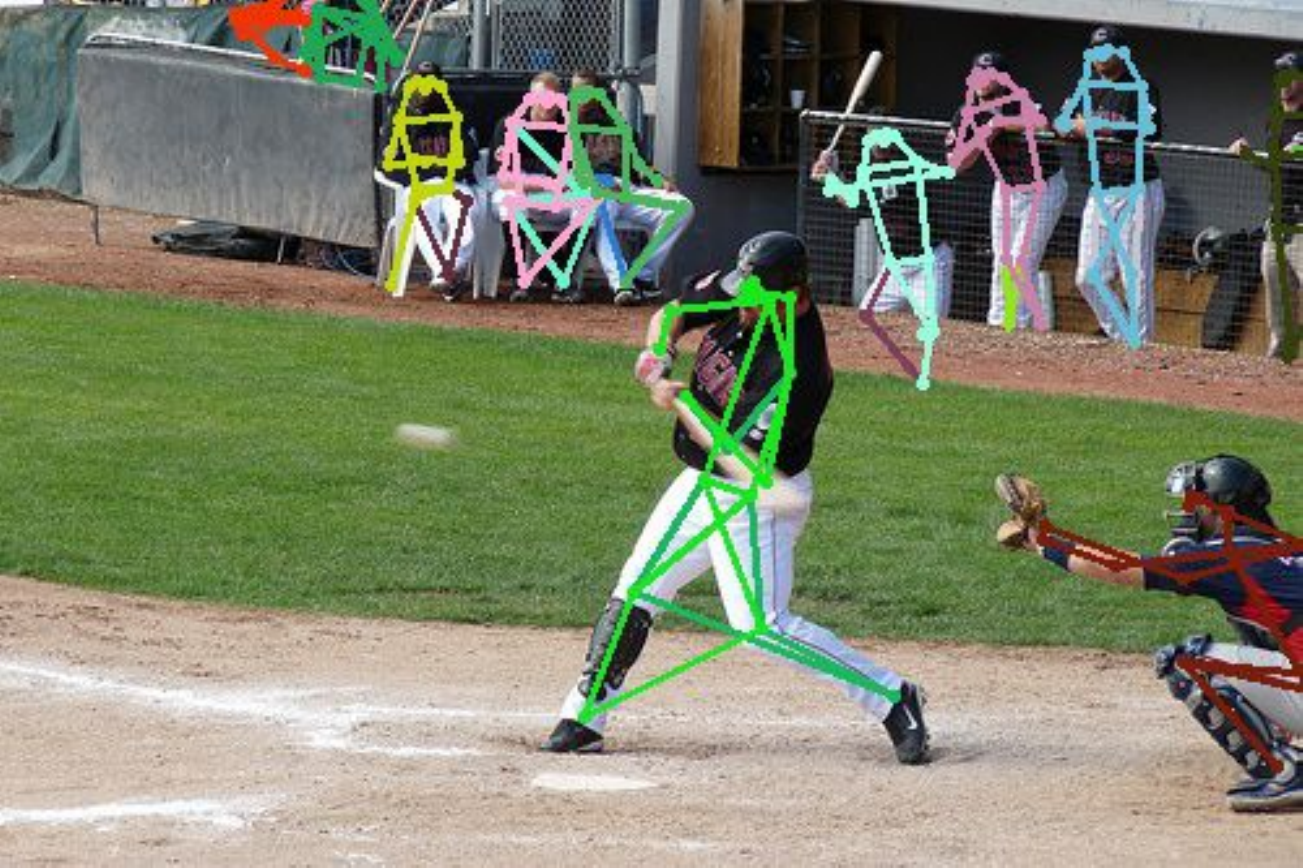}
    \end{minipage}  
    \hfill
    \begin{minipage}[t]{0.15\textwidth}
        \centering
        \includegraphics[width=2.6cm, height=2.15cm]{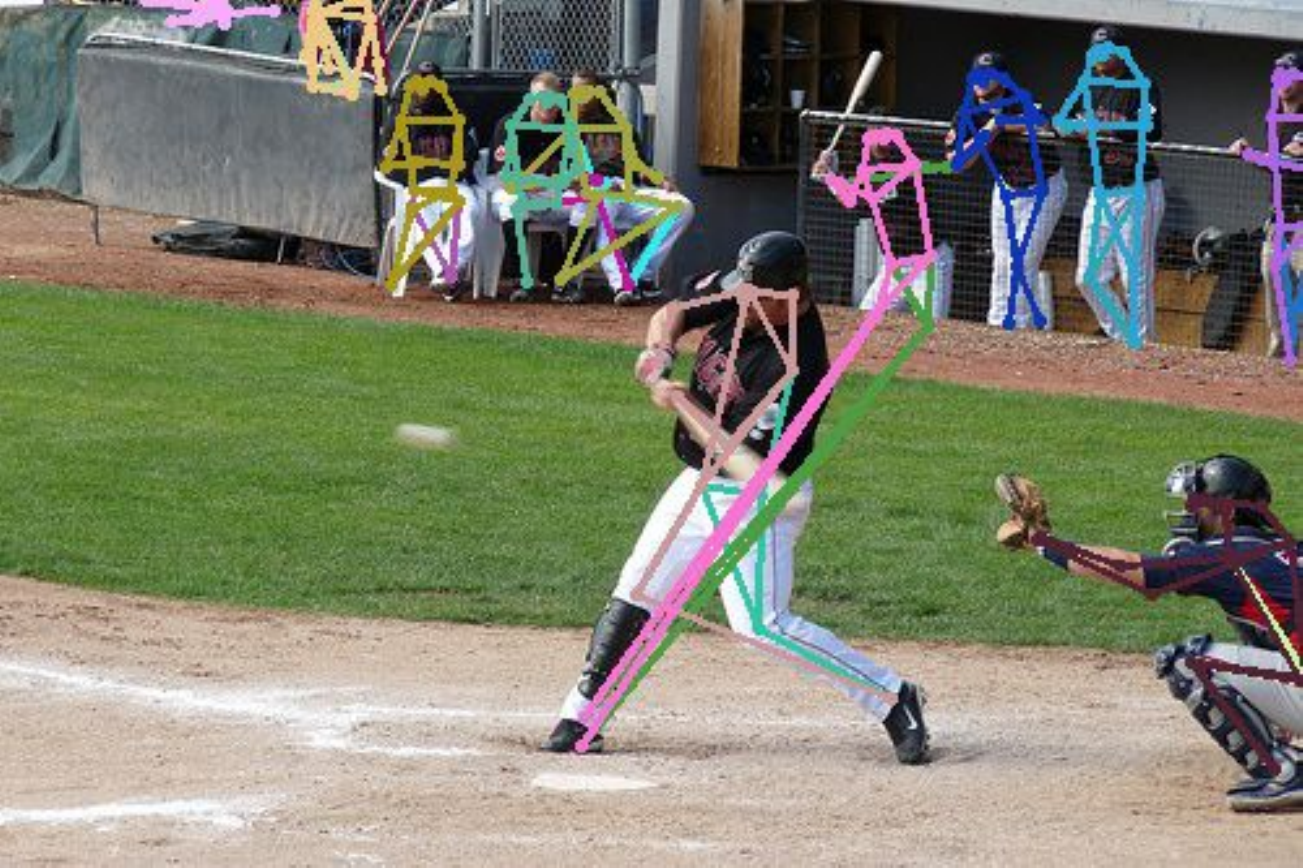}
    \end{minipage}
    \hfill
    \begin{minipage}[t]{0.15\textwidth}
        \centering
        \includegraphics[width=2.6cm, height=2.15cm]{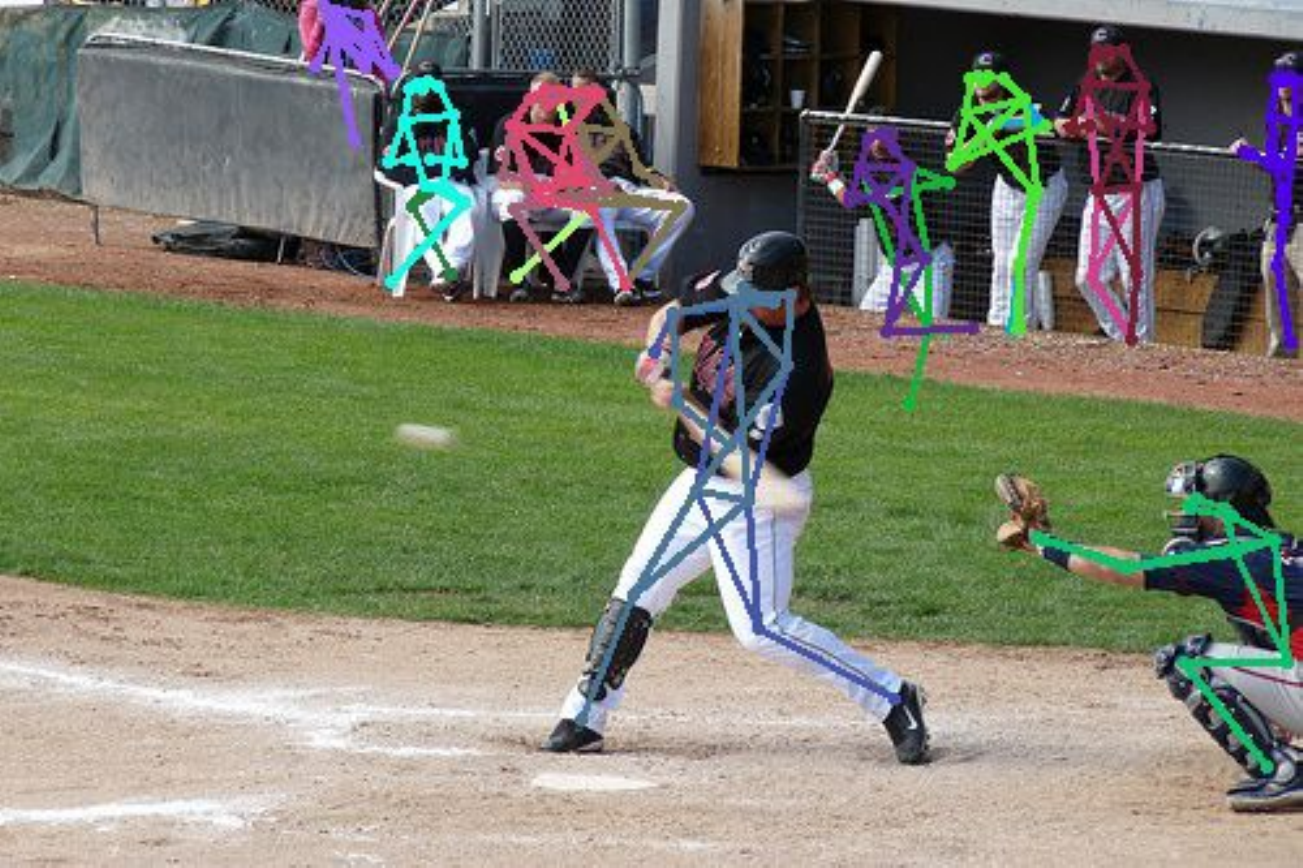}
    \end{minipage}  
    \hfill
    \caption{Qualitative Results for EfficientHRNet pose estimation models on COCO2017 test dataset. Top to bottom: Simple, Medium and Complex examples.}
    \label{fig:pose_qual}
\end{figure*}

%% file: tex/Conclusion.tex
\section{Conclusion} \label{sec:Conclusion}

In this paper, we have presented EfficientHRNet, a family of scalable networks for high-resolution and efficient bottom-up multi-person pose estimation made for real-time execution on low-power edge devices. EfficientHRNet unifies the principles of state-of-the-art EfficientNet \cite{efficientnet} and HRNet \cite{hrnet_pose} to create a network architecture for lightweight real-time human pose estimation, and proposes a new compound scaling method that jointly scales down the input resolution, backbone network, and high-resolution feature network. EfficientHRNet is not only more efficient than all other bottom-up human pose estimation methods, but it can maintain accuracy competitive with state-of-the-art models on the challenging COCO dataset. Remarkably, EfficientHRNet can achieve this near state-of-the-art accuracy with fewer parameters and less computational complexity than other bottom-up multi-person pose estimation networks, all while being able to achieve 23 FPS on an Nvidia Jetson NX Xavier.



%% file: tex/Acknowlegment.tex
\section*{Acknowledgement}

This research is supported by the National Science Foundation (NSF) under Awards No. 1831795.